\documentclass[sigconf]{acmart}

\setcopyright{none}
\renewcommand\footnotetextcopyrightpermission[1]{}
\AtBeginDocument{%
  \providecommand\BibTeX{{%
    \normalfont B\kern-0.5em{\scshape i\kern-0.25em b}\kern-0.8em\TeX}}}

\setcopyright{acmcopyright}
\copyrightyear{2021}
\acmYear{2021}
\acmConference[WWW '21]{WWW '21: The Web Conference}{April 19--23, 2021}{Ljubljana, Slovenia}
\acmBooktitle{WWW '21: The Web Conference,
  April 19--23, Ljubljana, Slovenia}
\acmPrice{15.00}
\acmISBN{978-1-4503-XXXX-X/18/06}

\usepackage{array,tabularx,makecell,multirow,amsmath,graphicx,algorithm,algpseudocode,epstopdf}

\newcolumntype{G}{>{\raggedright\arraybackslash\hsize=.82\hsize\linewidth=\hsize}X}
\newcolumntype{P}{>{\raggedright\arraybackslash\hsize=.82\hsize\linewidth=\hsize}X}
\newcolumntype{Q}{>{\raggedright\arraybackslash\hsize=.36\hsize\linewidth=\hsize}X}
\newcolumntype{S}{>{\centering\arraybackslash}m{0.085\textwidth}}

\usepackage{CJKutf8}
\usepackage[utf8]{inputenc}
\usepackage[T1]{fontenc}

\begin{document}

\title{Enquire One's Parent and Child Before Decision: Fully Exploit Hierarchical Structure for Self-Supervised Taxonomy Expansion}


% ORIGINAL AUTHOR LIST
\author{Suyuchen Wang}
\email{suyuchen.wang@umontreal.ca}
\orcid{0000-0003-0404-2921}
\affiliation{
  \institution{Mila \& DIRO, Universit{\'e} de Montr{\'e}al}
  \city{Montr{\'e}al}
  \state{Qu{\'e}bec}
  \country{Canada}
}
\authornote{Work done during an internship at Tencent Jarvis Lab.}

\author{Ruihui Zhao}
\email{zacharyzhao@tencent.com}
\affiliation{
  \institution{Tencent Jarvis Lab}
  \city{Shenzhen}
  \state{Guangdong}
  \country{China}
 }
 
\author{Xi Chen}
\email{jasonxchen@tencent.com}
\affiliation{
  \institution{Tencent Jarvis Lab}
  \city{Shenzhen}
  \state{Guangdong}
  \country{China}
 }
  
\author{Yefeng Zheng}
\email{yefengzheng@tencent.com}
\affiliation{
  \institution{Tencent Jarvis Lab}
  \city{Shenzhen}
  \state{Guangdong}
  \country{China}
}

\author{Bang Liu}
\email{bang.liu@umontreal.ca}
\affiliation{
  \institution{Mila \& DIRO, Universit{\'e} de Montr{\'e}al}
  \city{Montr{\'e}al}
  \state{Qu{\'e}bec}
  \country{Canada}
}
\authornote{Corresponding author.}

% ANOTHER AUTHOR LIST
% \author{Suyuchen Wang$^{1*}$, Bang liu$^{1\dagger}$, Ruihui Zhao$^2$, Xi Chen$^2$, Yefeng Zheng$^2$}
% \affiliation{$^1$Universit{\'e} de Montr{\'e}al, Montr{\'e}al, Qu{\'e}bec, Canada}
% \affiliation{$^2$Tencent Jarvis Lab, Shenzhen, Guangdong, China}
% \thanks{$^*$Work done during an internship at Tencent Jarvis Lab.}
% \thanks{$^\dagger$Correnspondence to: bang.liu@umontreal.ca.}

\renewcommand{\shortauthors}{Wang, et al.}

\begin{abstract}

Taxonomy is a hierarchically structured knowledge graph that plays a crucial role in machine intelligence.
The taxonomy expansion task aims to find a position for a new term in an existing taxonomy to 
capture the emerging knowledge in the world and 
keep the taxonomy dynamically updated. Previous taxonomy expansion solutions neglect valuable information brought by the hierarchical structure and evaluate the correctness of merely an added edge, which downgrade the problem to node-pair scoring or mini-path classification. In this paper, we propose the Hierarchy Expansion Framework (\textsf{HEF}), which fully exploits the hierarchical structure's properties to maximize the coherence of expanded taxonomy. \textsf{HEF} makes use of taxonomy's hierarchical structure in multiple aspects: \romannumeral1 ) \textsf{HEF} utilizes subtrees containing most relevant nodes as self-supervision data for a complete comparison of parental and sibling relations; \romannumeral2 ) \textsf{HEF} adopts a coherence modeling module to evaluate the coherence of a taxonomy's subtree by integrating hypernymy relation detection and several tree-exclusive features; \romannumeral3 ) \textsf{HEF} introduces the Fitting Score for position selection, which explicitly evaluates both path and level selections and takes full advantage of parental relations to interchange information for disambiguation and self-correction. Extensive experiments show that by better exploiting the hierarchical structure and optimizing taxonomy's coherence, \textsf{HEF} vastly surpasses the prior state-of-the-art on three benchmark datasets by an average improvement of 46.7\% in accuracy and 32.3\% in mean reciprocal rank.

\end{abstract}




\maketitle
\pagestyle{plain}
%%
%% Main texts

\section{Introduction}

\begin{figure}[!t]
  \centering
  \includegraphics[width=\linewidth]{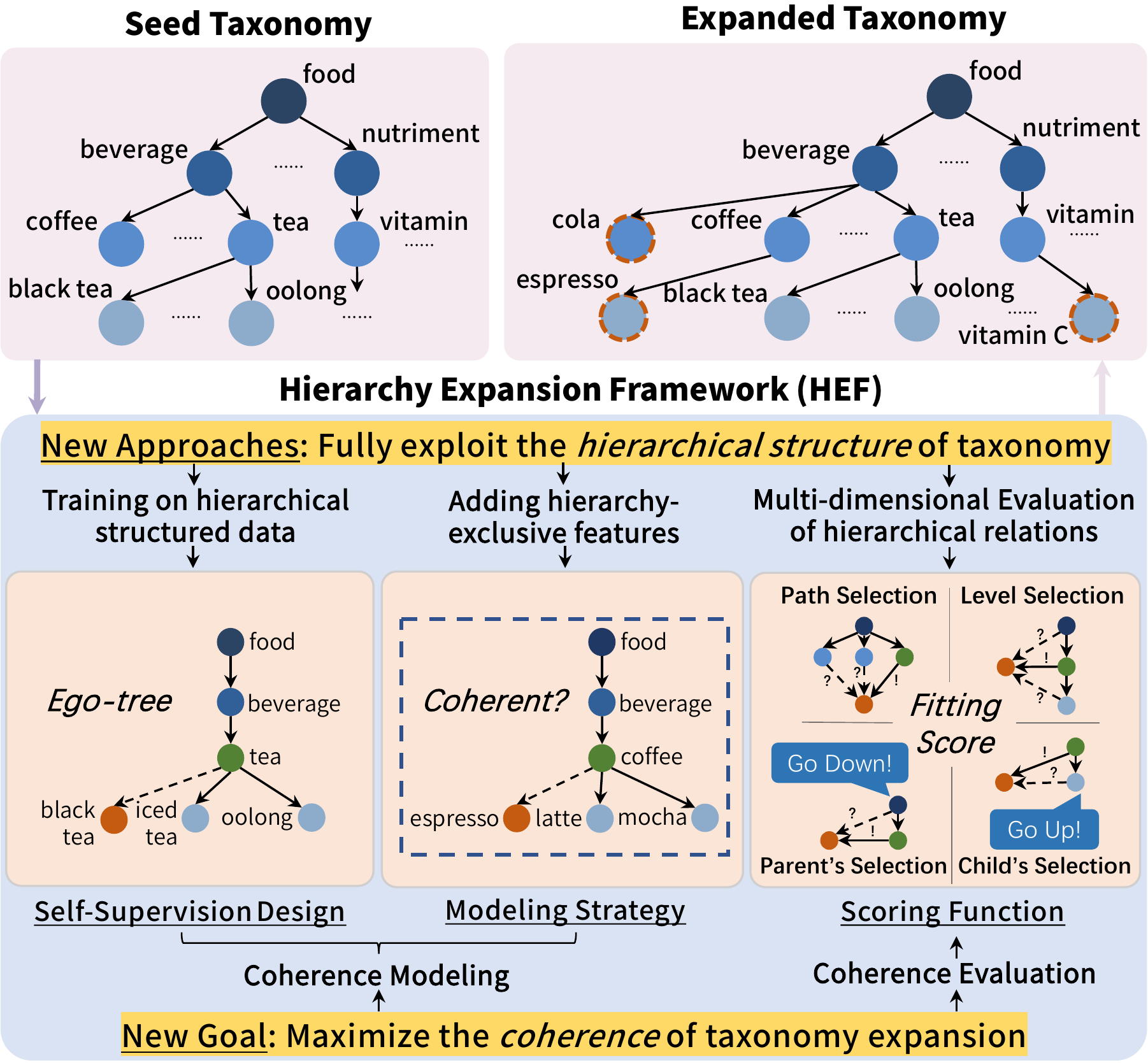}
  \caption{An illustration of the taxonomy expansion task and the contributions of the proposed \textsf{HEF} model.}
  \label{fig:task_example}
  \vspace{-3mm}
\end{figure}

Taxonomy is a particular type of hierarchical knowledge graph that portrays the hypernym-hyponym relations or {\em ``is-A''} relations of various concepts and entities. They have been adopted as the underlying infrastructure of a wide range of online services in various domains, such as product catalogs for e-commerce \cite{karamanolakis_txtract_2020, luo_alicoco_2020}, scientific indices like {\em MeSH} \cite{lipscomb_medical_2000}, and lexical databases like {\em WordNet} \cite{miller_wordnet_1995}. A well-constructed taxonomy can assist various downstream tasks, including web content tagging \cite{liu_user-centered_2019, peng_hierarchical_2019}, web searching \cite{yin_building_2010}, personalized recommendation \cite{huang_taxonomy-aware_2019} and helping users achieve quick navigation on web applications \cite{hua_understand_2017}.
Manually constructing and maintaining a taxonomy is laborious, expensive and time-consuming. It is also highly inefficient and detrimental for downstream tasks if we construct a taxonomy {\em from scratch} \cite{velardi_ontolearn_2013, gupta_taxonomy_2017} as long as the taxonomy has new terms to be added.
A more realistic strategy is to insert new terms ({\em ``query''}) into an existing taxonomy, i.e., the seed taxonomy, as a child of an existing node in the taxonomy ({\em ``anchor''}) without modifying its original structure to best preserve its design. This problem is called {\em taxonomy expansion} \cite{jurgens_semeval-2016_2016}.

Early taxonomy expansion approaches use terms that do not exist in the seed taxonomy and its best-suited position in the seed taxonomy as training data \cite{jurgens_reserating_2015}. However, it suffers from the insufficiency of training data and the shortage of taxonomy structure supervision. More recent solutions adopt self-supervision and try to exploit the information of nodes in the seed taxonomy (seed nodes) to perform node pair matching \cite{shen_taxoexpan_2020} or classification along mini paths in the taxonomy \cite{yu_steam_2020}.
% The construction and maintenance of early taxonomies are mainly completed by human experts, which suffers from the problem of {\em low coverage}, since it is laborious and expensive and lacks timeliness, especially when dealing with a large, growing taxonomy like MeSH\cite{lipscomb_medical_2000}. Hence, more recent approaches manage to automatically construct taxonomies, usually by first detect hypernymy relations among the terms \cite{lin_information-theoretic_1998, dash_hypernym_2020} and then prune the generated hypernym-hyponym pairs into a tree \cite{velardi_ontolearn_2013, gupta_taxonomy_2017}. However, the majority of these solutions can only building a taxonomy {\em from scratch}, while in reality, taxonomies ought to preserve underlying design outlined by human experts. Besides, generating a new taxonomy without keeping its original structure is impractical, for it is both time-consuming and detrimental for downstream tasks due to its instability.
However, these approaches do not fully utilize the taxonomy's hierarchical structure's characteristics, and neglect the coherence of the extended taxonomy which oughts to be the core of the taxonomy expansion task.
More specifically, existing approaches do not model a hierarchical structure identical to the taxonomy. Instead, they use ego-nets \cite{shen_taxoexpan_2020} or mini-paths \cite{yu_steam_2020} and feature few or no tree-exclusive information, making them unable to extract or learn the complete hierarchical design of a taxonomy. Besides, they do not consider the coherence of a taxonomy. They manage to find the most suitable node in a limited subgraph and only evaluate the correctness of a single edge instead of the expanded taxonomy, which downgrades the taxonomy expansion task to a hypernymy detection task. Lastly, their scoring approach regards the anchor node as an individual node without considering the hierarchical context information. However, the hierarchical structure provides multi-aspect criteria to evaluate a node, such as its path or level correctness. The structure also marks the nodes that are most likely to be wrongly chosen to be a parent in a specific parental relation.

% However, these prior arts does not fully utilize the characteristics of taxonomies' hierarchical structure. Firstly, these methods simply generalize the tree-structured taxonomy to a graph, neglecting important tree-exclusive information such as the similarity of siblings, parental relations and node depths, all of which potentially contains information about the construction of a taxonomy. Secondly, since a node's ancestors are all its hypernyms, matching scores considering only a node pair or a mini-path is insufficient for selecting the most suitable parent. Thirdly, the rule-based lexical-syntactic features and the usage of external corpus in these methods are coarsely designed, which not only hurts the performance by not offering insufficient knowledge of a term, but also brings noise since co-occurrence does not guarantee hypernym-hyponym relations. In one word, the major flaw of these methods is that they actually downgrades the taxonomy expansion task as a special kind of hypernymy detection task to predict whether a seed node is the parent of a query term, but never considers the coherence of taxonomy design after adding a new term, which requires evaluation of a {\em tree structure} rather than node pairs, and explicit disambiguation by comparing the hypernyms of a query term.

To solve all the stated flaws in previous works, we propose the Hierarchy Expansion Framework (\textsf{HEF}), which aims to maximize the coherence of the expanded taxonomy instead of the fitness of a single edge by fully exploiting the hierarchical structure of a taxonomy for self-supervised training, as well as modeling and evaluating the structure of taxonomy. \textsf{HEF}'s designs and goals are illustrated in Fig.~\ref{fig:task_example}. Specifically, we make the following contributions. 

Firstly, we design an innovative hierarchical data structure for self-supervision to mimic how humans construct a taxonomy. Relations in a taxonomy include hypernymy relations along a root-path and similarity among siblings. To find the most suitable parent node for the query term, human experts need to compare an anchor node with all its ancestors to distinguish the most appropriate one and compare the query with its potential siblings to testify their similarity. For example, to choose the parent for query {\em ``black tea''} in the food taxonomy, the most appropriate anchor {\em ``tea''} can only be selected by distinguishing from its ancestors {\em ``beverage''} and {\em ``food''}, which are all {\em ``black tea''} 's hypernyms, as well as compare the query {\em ``black tea''} with {\em ``tea''}'s children like {\em ``iced tea''} and {\em ``oolong''} to guarantee similarity among siblings. Thus, we design a new structure called {\em ``ego-tree''} for self-supervision, which contains all ancestors and sample of children of a node for taxonomy structure learning. Our ego-tree incorporates richer topological context information for attaching a query term to a candidate parent with minimal computation cost compared to previous approaches based on node pair matching or path information.

% Firstly, we model a complete tree structure \red{to mimic the way how human constructs a taxonomy}. \textbf{\red{NOTE: explain the principle how a human construct a taxonomy.}} We design a new structure called "{\em ego-tree}" for self-supervision, which contains all ancestors and sample of children of a node for taxonomy structure learning. Compared to previous approaches based on node pair matching or path information, our ego-tree incorporates richer topological context information when attaching a new item to a candidate parent node, enabling the model to \red{learn the complete construction of a taxonomy} while minimizing the computation cost.

Secondly, we design a new modeling strategy to perform explicit ego-tree coherence modeling apart from the traditional node-pair hypernymy detection. Instead of merely modeling the correctness of the added edge, we adopt a more comprehensive approach to detect whether the anchor's ego-tree after adding the query maintains the original design of the seed taxonomy. The design of taxonomy includes natural hypernymy relations, which needs the representation of node-pair relations and expert-curated level configurations, such as {\em species} must be placed in the eighth level of biological taxonomy, or adding one more adjective to a term means exactly one level higher in the e-commerce taxonomy. We adopt a coherence modeling module to detect the two aspects of coherence: \romannumeral1 ) For natural hypernymy relations, we adopt a hypernymy detection module to represent the relation between the query and each node in the anchor's ego-tree. \romannumeral2 ) For expert-curated designs, we integrate hierarchy-exclusive features such as embeddings of a node's absolute level and relative level to the anchor into the coherence modeling module.

% Secondly, apart from the hypernymy detection module for node-wise hypernymy detection, we explicitly design a coherence modeling module for modeling ego-tree coherence. Instead of modeling with anchor and query node's representation in different vector spaces to evaluate the fitness of adding an edge, we adopt a more comprehensive and homogeneous approach by trailing the query as a node in taxonomy and evaluate the cohesiveness of the whole structure. Moreover, to assist learning the intrinsic design of a taxonomy such as granularity configuration on different levels, we add several tree-exclusive features to assist the modeling of ego-trees, such as a node's absolute and relative level to the anchor. These features allow the model to evaluate the structure coherence in different aspects.

% Moreover, for the model to better understand the intrinsic design of a taxonomy such as granularity configuration on different level, we also add several tree-exclusive features like absolute and relative level of nodes to the model.

Thirdly, we design a multi-dimensional evaluation to score the coherence of the expanded taxonomy. The hierarchical structure of taxonomy allows the model to evaluate the correctness of path selection and level selection separately and the parental relationships in a hierarchy not only allow the model to disambiguate the most similar terms but also enables the model to self-correct its level selection by deciding the current anchor's granularity is too high or too low. We introduce the Fitting Score for the coherence evaluation of the expanded ego-tree by using a Pathfinder and a Stopper to score path correctness and level correctness, respectively. The Fitting Score calculation also disambiguates the most appropriate anchor from its parent and children and self-correct its level selection by bringing the level suggestion from the anchor's parent and one of its children into consideration. The Fitting Score's optimization adopts a self-supervised multi-task training paradigm for the Pathfinder and Stopper, which automatically generates training data from the seed taxonomy to utilize its information fully.

We conduct extensive evaluations based on three benchmark datasets to compare our method with state-of-the-art baseline approaches. The results suggest that the proposed \textsf{HEF} model significantly surpasses the previous solutions on all three datasets by an average improvement of 46.7\% in accuracy and 32.3\% in mean reciprocal rank. A series of ablation studies further demonstrate that \textsf{HEF} can effectively perform the taxonomy expansion task.

\section{Related Work}

\textbf{Taxonomy Construction}.
Taxonomy construction aims to create a tree-structured taxonomy with a set of terms (such as concepts and entities) from scratch, integrating hypernymy discovery and tree structure alignment. It can be further separated into two subdivisions. The first focuses on topic-based taxonomy, where each node is a cluster of several terms sharing the same topic \cite{zhang_taxogen_2018, shang_nettaxo_2020}. The other subdivision tackles the problem of term-based taxonomy construction, in which each node represents the term itself \cite{cocos_comparing_2018, shen_hiexpan_2018, mao_end--end_2018}. A typical pipeline for this task is to extract {\em ``is-A''} relations with a hypernymy detection model first using either a pattern-based model \cite{hearst_automatic_1992, agichtein_snowball_2000, jiang_metapad_2017, roller_hearst_2018} or a distributional model \cite{lin_information-theoretic_1998, yin_term_2018, wang_family_2019, dash_hypernym_2020}, then integrate and prune the mined hypernym-hyponym pairs into a single directed acyclic graph (DAG) or tree \cite{gupta_taxonomy_2017}. More recent solutions utilize hyperbolic embeddings \cite{le_inferring_2019} or transfer learning \cite{shang_taxonomy_2020} to boost performance.

\textbf{Taxonomy Expansion}.
In the taxonomy expansion task, an expert-curated seed taxonomy like MeSH \cite{lipscomb_medical_2000} is provided as both the guidance and the base for adding new terms. The taxonomy expansion task is a ranking task to maximize a score of a node and its ground-truth parent in the taxonomy. 
Wang et al. \cite{wang_hierarchical_2014} adopted Dirichlet distribution to model the parental relations. 
ETF \cite{vedula_enriching_2018} trained a learning-to-rank framework with handcrafted structural and semantic features. 
% Aly et al. \cite{aly_every_2019} introduced a refinement algorithm based on hyperbolic embeddings. 
Arborist \cite{manzoor_expanding_2020} calculated the ranking score in a bi-linear form and adopted margin ranking loss. TaxoExpan \cite{shen_taxoexpan_2020} modeled the anchor node by passing messages from its egonet instead of considering a single node, and scored by feeding a concatenation of egonet representation and query embedding to a feed-forward layer. STEAM \cite{yu_steam_2020} transformed the scoring task into a classification task on mini-paths and performed model ensemble of three sub-models processing distributional, contextual, and lexical-syntactic features, respectively. However, existing approaches mostly neglect the characteristics of taxonomy's hierarchical structure and only evaluate the correctness of a single edge from anchor to query. On the contrary, our method utilizes the features and relations brought by the hierarchical structure and aims to enhance the expanded taxonomy's overall coherence.

\textbf{Modeling of Tree-Structured Data}.
Taxonomy expansion involves modeling a tree or graph structure. Plenty of works have been devoted to extending recurrent models to tree structures, like Tree-LSTM \cite{tai_improved_2015}. For explicit tree-structure modeling, previous approaches include modeling the likelihood of a Bayesian network \cite{fountain_taxonomy_2012, wang_hierarchical_2014} or using graph neural net variants \cite{shen_taxoexpan_2020, yu_steam_2020}. Recently, Transformers \cite{vaswani_attention_2017} achieved state-of-the-art performance in the program translation task by designing a novel positional encoding related to paths in the tree \cite{shiv_novel_2019} or merely transforming a tree to sequence by traversing its nodes \cite{kim_code_2020}. In our work, we model tree-structure by a Transformer encoder, which, to the best of our knowledge, is the first to use the Transformer for taxonomy modeling. We adopt a more natural setting than \cite{shiv_novel_2019} by using two different embeddings for a node's absolute and relative level to denote positions.

\begin{figure*}[!t]
  \centering
  \includegraphics[width=\textwidth]{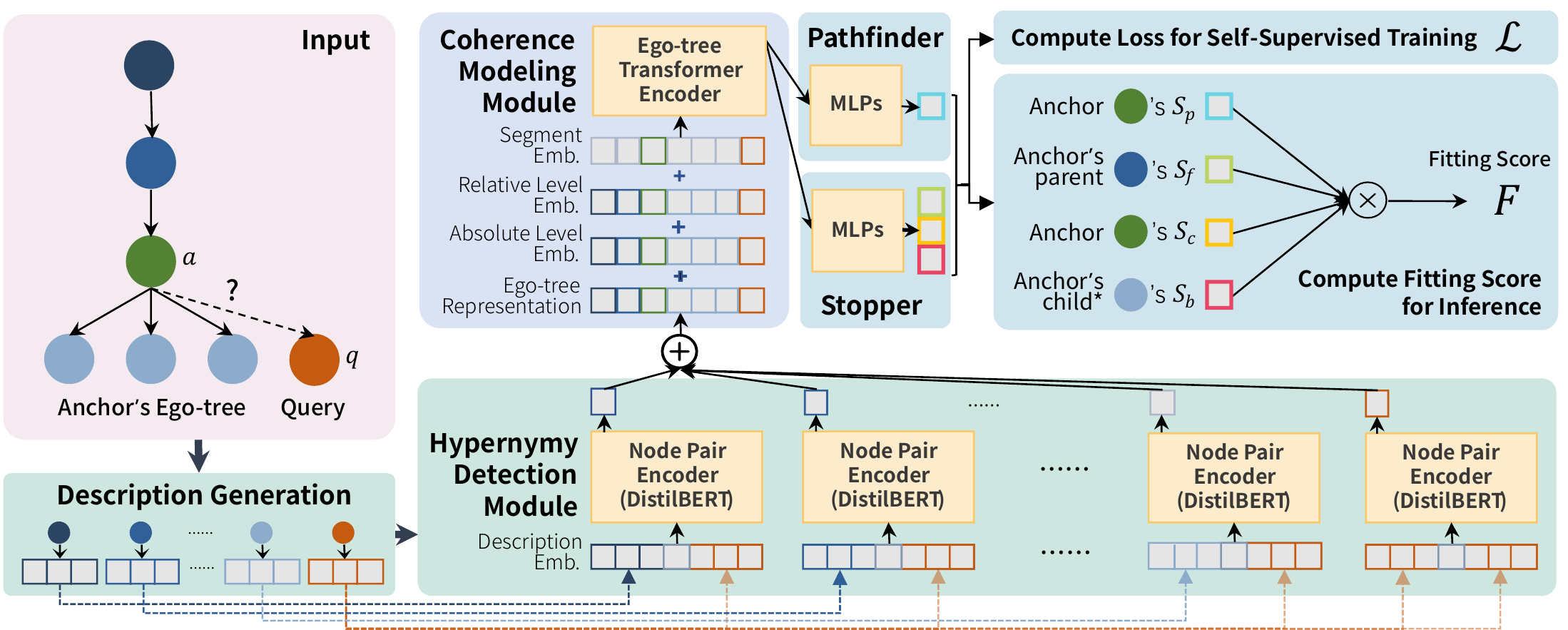}
  \caption{Illustration of the \textsf{HEF} model. Each circle denotes a seed node or a query node. The ``Anchor's child*'' in Fitting Score calculation denotes the anchor's child with maximum Pathfinder Score $S_p$.}
  \label{fig:model}
\end{figure*}

\section{Problem Definition}

In this section, we provide the formal definition of the taxonomy expansion task and the explanation of key concepts that will occur in the following sections.

\textbf{Definition and Concepts about Taxonomy.} A taxonomy $\mathcal{T}=(\mathcal{N},\mathcal{E})$ is an arborescence that presents hypernymy relations among a set of nodes. Each node $n\in\mathcal{N}$ represents a {\em term}, usually a concept mined from a large corpus online or an artificially extracted phrase. Each edge $\left<n_p, n_c\right>\in\mathcal{E}$ points to a node from its most exact hypernym node, where $n_p$ is $n_c$'s {\em parent} node, and $n_c$ is $n_p$'s {\em child} node. Since hypernymy relation is transitive \cite{sang_extracting_2007}, such relation exists not only in node pairs connected by a single edge, but also in node pairs connected by a path in the taxonomy. Thus, for a node $n$ in the taxonomy, its hypernym set and hyponym set consists of its {\em ancestors} $\mathcal{A}_n$, and its {\em descendants} $\mathcal{D}_n$ respectively.

\textbf{Definition of the Taxonomy Expansion Task.} Given a {\em seed taxonomy} $\mathcal{T}^0=(\mathcal{N}^0,\mathcal{E}^0)$ and the set of terms $\mathcal{C}$ to be added to the seed taxonomy, The model outputs the taxonomy $\mathcal{T}=(\mathcal{N}^0\cup\mathcal{C},\mathcal{E}^0\cup\mathcal{R})$, where $\mathcal{R}$ is the newly added relations from seed nodes in $\mathcal{N}^0$ to new terms in $\mathcal{C}$. More specifically, during the inference phase of a taxonomy expansion model, when given a {\em query} node $q\in\mathcal{C}$, the model finds its best-suited parent node by iterating each node in the seed taxonomy as an {\em anchor} node $a\in\mathcal{N}^0$, calculating a score $f(a,q)$ representing the suitability for adding the edge $\left<a,q\right>$, and deciding $q$'s parent $p_q$ in the taxonomy by $p_q=\mathop{\arg\max}_{a\in\mathcal{N}^0}{f(a,q)}$.

\textbf{Accessible External Resources.} As a term's surface name is usually insufficient to convey the semantic information for hypernymy relationship detection, previous research usually utilizes term definitions \cite{jurgens_semeval-2016_2016, shen_taxoexpan_2020} or related web pages \cite{wang_hierarchical_2014, kozareva_semi-supervised_2010} to learn term representations. Besides, existing hypernymy detection solutions usually use large external corpora to discover lexical or syntactic patterns \cite{shwartz_improving_2016, yu_steam_2020}. As for the SemEval-2016 Task 13 datasets \cite{bordea_semeval-2016_2016} used for our model's evaluation, utilizing the WordNet \cite{miller_wordnet_1995} definitions is allowed by the original task, which guarantees a fair comparison with previous solutions.

\section{The Hierarchy Expansion Framework}

In this section, we introduce the design of the Hierarchy Expansion Framework (\textsf{HEF}). An illustration of \textsf{HEF} is shown in Fig.~\ref{fig:model}. We first introduce the way \textsf{HEF} models the coherence of a tree structure, including two components for node pair hypernymy detection and ego-tree coherence modeling, respectively. Then, we discuss how \textsf{HEF} further exploits the hierarchical structure for multi-dimensional evaluation by the modules of Pathfinder and Stopper, and the self-supervised paradigm to train the model for the Fitting Score calculation.

\subsection{Node Pair Hypernymy Detection}
\label{subsec:hypernymydetect}

We first introduce the hypernymy detection module of \textsf{HEF}, which detects the hypernymy relationships between two terms. Unlike previous approaches that manually design a set of classical lexical-syntactic features, we accomplish the task more directly and automatically by expanding the surface names of terms to their descriptions and utilizing pre-trained language models to represent the relationship between two terms.

Given a seed term $n\in \mathcal{N}^0$ and a query term $q\in \mathcal{N}^0$ during training or $q\in \mathcal{C}$ during inference, the hypernymy detection module outputs a representation $r_{n,q}$ suggesting how well these two terms form a hypernymy relation. Note that $n$ might not be identical to the anchor $a$. Since the surface names of terms do not contain sufficient information for relation detection, we expand the surface names to their descriptions, enabling the model to better understand the semantic of new terms. We utilize the WordNet \cite{miller_wordnet_1995} concept definitions for completing this task. However, WordNet cannot explain all terms in a taxonomy due to its low coverage. Besides, many terms used in taxonomies are complex phrases like {\em ``adaptation to climate change''} or {\em ``bacon lettuce tomato sandwich''}. Therefore, we further develop a description generation algorithm \texttt{descr($\cdot$)}, which generates meaningful and domain-related descriptions for a given term based on WordNet. Specifically, \texttt{descr($\cdot$)} is a dynamic programming algorithm that tends to integrate tokens into longer and explainable noun phrases. It describes each noun phrase by the most relative description to the taxonomy's root's surface name for domain relevance. The details are shown in Alg. \ref{alg:description} in the appendix. The input for hypernymy detection is organized as the input format of a Transformer:
\begin{displaymath}
  D_{n,q}=\left[\mbox{<CLS>}\oplus\texttt{descr(}n\texttt{)}\oplus\mbox{<SEP>}\oplus\texttt{descr(}q\texttt{)}\oplus\mbox{<SEP>}\right],
\end{displaymath}
where $\oplus$ represents concatenation, {\em <CLS>} and {\em <SEP>} are the special token for classification and sentence separation in the Transformer architecture, respectively. 

As shown in Fig.~\ref{fig:model}, the hypernymy detection module utilizes a pre-trained DistilBERT \cite{sanh_distilbert_2020}, a lightweight variant of BERT \cite{devlin_bert_2019}, to learn the representations of cross-text relationships.
Specifically, we first encode $D_{n,q}$ by ${\rm DistilBERT}(\cdot)$ with positional encoding.
%Following common practice of using BERT-based model for classification tasks,
Then we take the final layer representation of {\em <CLS>} as the representation of the node pair $\left<n,q\right>$: 
\begin{displaymath}
  r_{n,q}={\rm DistilBERT}\left(D_{n,q}\right)\left[0\right],
\end{displaymath}
where index $0$ represents the position of {\em <CLS>}'s embedding.

\subsection{Ego-Tree Coherence Modeling}

% \blue{Although the hypernymy detection module helps to learn whether a candidate anchor node $a$ is a parent of a query term $q$, it is not enough to select the most appropriate parent node for $q$ in the taxonomy. Because any ancestor node of $a$ can also be a parent of $q$.}
% \blue{The hypernymy detection module is insufficient for choosing the most appropriate anchor node. Since a node's ancestors in the taxonomy all hold hypernymy relations with it, an explicit comparison among a node's ancestors is needed to distinguish the most appropriate one. This step is to maintain the coherence of the whole taxonomy ins.}
% Thus, we further design a coherence modeling module to evaluate the coherence of the tree structure after attaching the query term $q$ into taxonomy $\mathcal{T}$ as the anchor $a\in\mathcal{N}^0$'s child.
% The coherence modeling module needs to: i) model a subtree of the taxonomy rather than a single node, enabling the model to \red{learn the construction of a hierarchy (What does this means????)}; ii) add tree-exclusive features to assist learning the latent \red{design (???representation?)} of the taxonomy; and iii) adopt a more homogeneous approach by trialling the query into the hierarchy as an existing node for coherence detection, instead of using representations in different vector space for structure modeling like \cite{shen_taxoexpan_2020,yu_steam_2020} \red{(iii) is hard to understand. Please revise it.)}.

We further design a coherence modeling module to evaluate the coherence of the tree structure after attaching the query term $q$ into taxonomy $\mathcal{T}$ as the anchor $a\in\mathcal{N}^0$'s child.

There are two different aspects for considering a taxonomy's coherence: \romannumeral1 ) the natural hypernymy relations. Since a node's ancestors in the taxonomy all hold hypernymy relations with it, an explicit comparison among a node's ancestors is needed to distinguish the most appropriate one; \romannumeral2 ) the expert-curated designs, which act as supplement information for maintaining the overall structure. Some taxonomies contain latent rules about a node's absolute or relative levels in a taxonomy. For example, in the biological taxonomy, {\em kingdoms} and {\em species} are all placed in the second and eighth levels, respectively; in some e-commerce catalog taxonomies, terms that are one level higher than another term contain exactly one more adjective.
Hence, the coherence modeling module needs to: \romannumeral1 ) model a subtree with the query as a node in it, rather than a single node pair, enabling the model to learn the design of a complete hierarchy; \romannumeral2 ) add tree-exclusive features like absolute level or relative level compared to the anchor to assist learning the expert-curated designs of the taxonomy.

We design the {\em Ego-tree} $\mathcal{H}_a$, a novel contextual structure of an anchor $a$, which consists of all the ancestors and children of $a$ (see Fig.~\ref{fig:model}). This structure contains all relevant nodes to both anchor and query, enabling the model to both compare all hypernymy relations along the root path and detect similarity among query and its potential siblings with minimal computation cost:
\begin{equation}
    \mathcal{H}_a=\mathcal{A}_a \cup \left\{a\right\} \cup {\rm sample\_child}\left(a\right),
\end{equation}
where $\mathcal{A}_a$ is all ancestors of $a$ in the seed taxonomy $\mathcal{T}^0$, and {\rm sample\_child}$(\cdot)$ means sampling at most three children of the anchor based on surface name similarity. The 3-children sampling is a trade-off between accuracy and speed, for three potential siblings are empirically enough for a comprehensive similarity comparison with the query (especially when these potential siblings are quite different) while decreasing the computation cost. Since this procedure is to leverage the similarity brought by a hierarchy's sibling relations, sampling by surface name similarity is intuitive and cost-saving given that similar surface names usually indicate similar terms.
The input of the coherence modeling module includes the anchor's ego-tree $\mathcal{H}_a$ and the query $q$ as the anchor's child in $\mathcal{H}_a$. For each node $n \in \mathcal{H}_a$, we represent the node pair $\left<n,q\right>$ by the following representations:
\begin{itemize}
    \item \textbf{Ego-tree representations.} The ego-tree representation $r_{n,q}$ is the output of the hypernymy detection module described in Sec.~\ref{subsec:hypernymydetect}. It suggests the node pair's relation.
    \item \textbf{Absolute level embedding.} The absolute level embedding $l_{n,q}=\mbox{AbsLvlEmb}\left(d_n\right)$, where $d_n$ is the depth of $n$ in the expanded taxonomy. When $n=q$, $l_{q,q}=\mbox{AbsLvlEmb}\left(d_a+1\right)$. It assists the modeling of the expert-curated designs about granularity of a certain level.
    \item \textbf{Relative level embedding.} The relative level embedding $e_{n,q}=\mbox{RelLvlEmb}\left(d_n-d_q\right)$, where $d_n$ is the depth of $n$ in the expanded taxonomy. It assists the modeling of expert-curated designs about the cross-level comparison.
    \item \textbf{Segment embedding.} The segment embedding of  $\left<n,q\right>$  $g_{n,q}=\mbox{SegEmb}\left(\mbox{segment}\left(n\right)\right)$ distinguishes anchor and query with other nodes in the ego-tree, where:
    \begin{displaymath}
      \mbox{segment}\left(n,q\right)=
      \begin{cases}
        0, & \mbox{if } n \mbox{ is the anchor}, \\
        1, & \mbox{if } n \mbox{ is the query}, \\
        2, & \mbox{otherwise}.
      \end{cases}
    \end{displaymath}
\end{itemize}

The input of the coherence modeling module $R_{a,q}\in\mathbb{R}^{\left(\mathopen|\mathcal{H}_a\mathclose|+3\right)\times d}$ is the sum of the above embeddings calculated with the anchor's ego-tree and the query, organized as the input of a Transformer:
\begin{equation}
    R_{a,q}=\left[e_{<CLS>}\oplus e_{<CLS>}\bigoplus_{n\in\mathcal{H}_a\cup{\left\{q\right\}}}{\left(r_{n,q}+l_{n,q}+e_{n,q}+g_{n,q}\right)}\right],
\end{equation}
where $d$ is the dimension of embedding, $e_{<CLS>}$ is a randomly initialized placeholder vector for obtaining the ego-tree's path and level coherence representations, and $\oplus$ denotes concatenation.

We implement the coherence modeling module using a Transformer encoder. Transformers are powerful to model sequences, but they lack positional information to process the relations among nodes in graphs. However, in a hierarchy like taxonomy, the level of nodes can be used as positional information, which simultaneously eliminates the positional difference of nodes on the same level.
%, solving the problem of using sequential models for hierarchies.
Transformers are also strong enough to integrate multiple-source information by adding their embeddings, thus they are quite suitable for modeling tree structures.
In our \textsf{HEF} model, as shown in Fig.~\ref{fig:model}, by using two {\em <CLS>}s in the module's input, we can obtain two different representations: $p_{a,q}$ representing the coherence of hypernymy relations (whether the {\em path} is correct), and $d_{a,q}$ representing the coherence of inter-level granularity (whether the {\em level} is correct), evaluating how well the query fits the current position in the taxonomy in both horizontal and vertical perspective:
\begin{align*}
  p_{a,q}&=\mbox{TransformerEncoder}\left(R_{a,q}\right)\left[0\right]\\
  d_{a,q}&=\mbox{TransformerEncoder}\left(R_{a,q}\right)\left[1\right],
\end{align*}
where $0$ and $1$ are the position indexes of the two {\em <CLS>}s.

\subsection{Fitting Score-based Training and Inference}

\begin{figure}[!t]
  \centering
  \includegraphics[width=\linewidth]{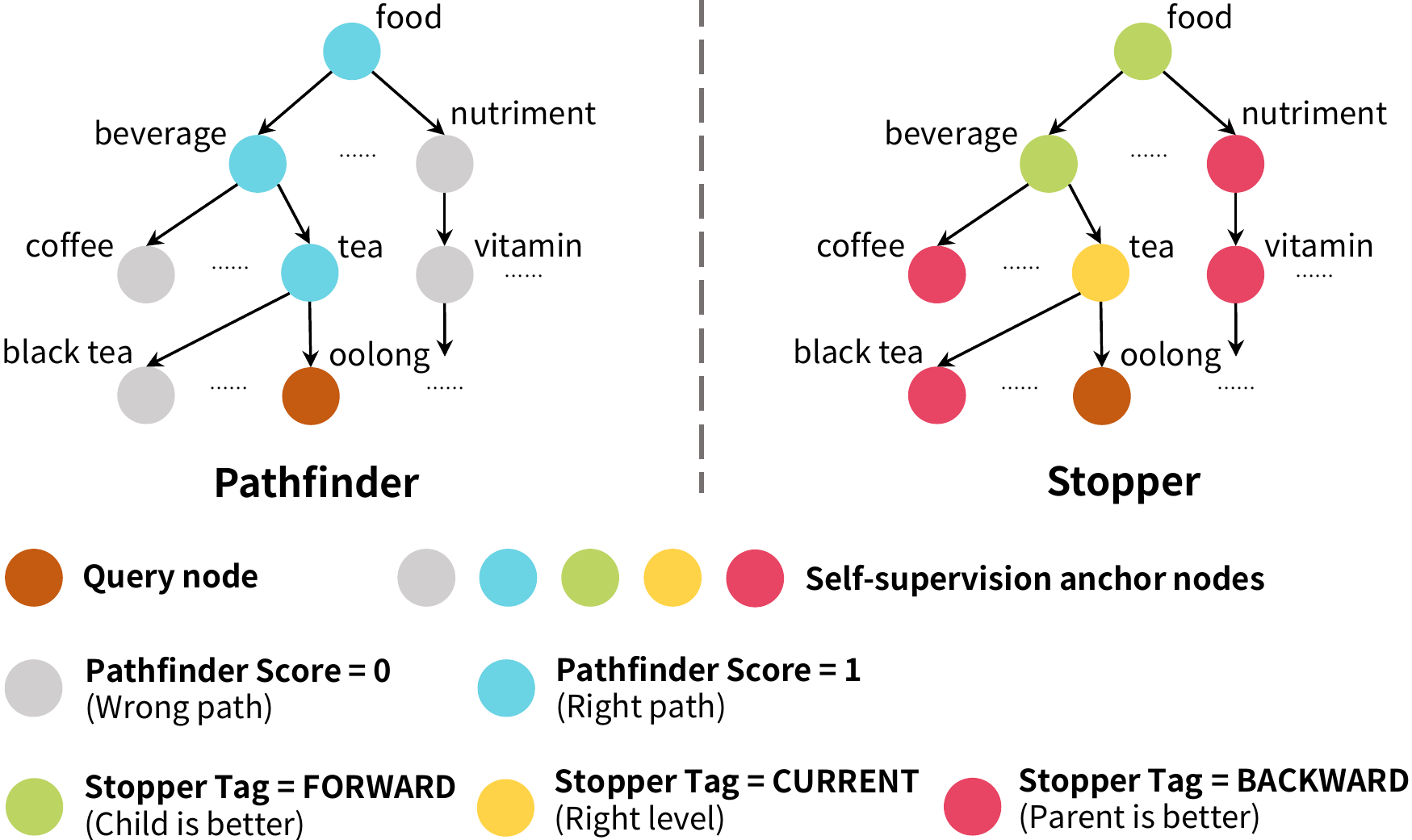}
  \caption{An illustration of the self-supervision data labels for Pathfinder and Stopper.}
  \label{fig:label}
\end{figure}

The two representations $p_{a,q}$ and $d_{a,q}$ need to be transformed into scores indicating the fitness of placing the query $q$ on a particular path and a particular level. Thus, we propose the {\em Pathfinder} for path selection and the {\em Stopper} for level selection, as well as a new self-supervised learning algorithm for training and the Fitting Score calculation for inference.

\textbf{Pathfinder.} The Pathfinder detects whether the query is positioned on the right path. This module performs a binary classification using $p_{a,q}$. The Pathfinder Score $S_p=1$ if and only if $a$ and $q$ are on the same root-path:
\begin{equation}
    \label{eqn:pathfinder}
    S_{p}\left(a,q\right)=\sigma\left(\mathbf{W}_{p2}\tanh\left(\mathbf{W}_{p1} p_{a,q}+b_{p1}\right)+b_{p2}\right),
\end{equation}
where $\sigma$ is the sigmoid function, and $\mathbf{W}_{p1},\mathbf{W}_{p2},b_{p1},b_{p2}$ are trainable parameters for multi-layer perceptrons.

\textbf{Stopper.} The Stopper detects whether the query $q$ is placed on the right level, i.e., under the most appropriate anchor $a$ on a particular path. Selecting the right level is nonidentical to selecting the right path since levels are kept in order. The order of nodes on a path enables us to design a more representative module for further classifying whether the current level is too high (anchor $a$ is a coarse-grained ancestor of $q$) or too low ($a$ is a descendant of $q$). Thus, the Stopper module uses $d_{a,q}$ to perform a $3$-class classification: searching for a better anchor node needs to go {\em Forward}, remain {\em Current}, or go {\em Backward}, in the taxonomy:
\begin{align}
    \label{eqn:stopper}
    \lefteqn{[S_f\left(a,q\right),S_c\left(a,q\right),S_b\left(a,q\right)]=}\nonumber\\[1ex]
    &\mbox{softmax}\left(\mathbf{W}_{s2}\tanh\left(\mathbf{W}_{s1} d_{a,q}+b_{s1}\right)+b_{s2}\right),
\end{align}
where $\mathbf{W}_{p1},\mathbf{W}_{p2},b_{p1},b_{p2}$ are trainable parameters for multi-layer perceptrons. {\em Forward Score} $S_f$, {\em Current Score} $S_c$ and {\em Backward Score} $S_b$ are called {\em Stopper Scores}.
% This multi-classification design allows us to perform disambiguation and self-correction by exchanging Stopper Scores with a node's parent and children for Fitting Score calculation.

\noindent\textbf{Self-Supervised Training.} Training the \textsf{HEF} model needs data labels for both the Pathfinder and the Stopper. The tagging scheme is illustrated in Fig.~\ref{fig:label}. There are totally four kinds of Pathfinder-Stopper label combinations since Pathfinder Score is always $1$ when Stopper Tag is {\em Forward} or {\em Current}. The training process of \textsf{HEF} is shown in Alg.~\ref{alg:training}. Specifically, we sample the ego-tree of all four types of nodes for a query: $q$'s parent $a$, $a$'s ancestors, $a$'s descendants and other nodes, as a mini-batch for training the Pathfinder and Stopper simultaneously.

The optimization of Pathfinder and Stopper can be regarded as a multi-task learning process. The loss $\mathcal{L}_q$ in Alg.~\ref{alg:training} is a linear combination of the loss from Pathfinder and Stopper:
\begin{align}
    \mathcal{L}_q=&-\eta\frac{1}{\mathopen|\mathcal{X}_q\mathclose|}\sum_{a\in\mathcal{X}_q}{{\rm BCELoss}\left(\hat{S_p}\left(a,q\right), S_p\left(a,q\right)\right)}\nonumber\\[1ex]
    &-\left(1-\eta\right)\frac{1}{\mathopen|\mathcal{X}_q\mathclose|}\sum_{a\in\mathcal{X}_q}{\sum_{k\in\left\{f,c,b\right\}}{\hat{s_k}\left(a,q\right)\log s_k\left(a,q\right)}},
    \label{eqn:loss}
\end{align}
where ${\rm BCELoss}\left(\cdot\right)$ denotes the binary cross entropy, and $\eta$ is the weight of multi-task learning.

\begin{algorithm}[t]
\caption{Self-Supervised Training Process of \textsf{HEF}.}
\label{alg:training}
\begin{algorithmic}[1]
\Procedure{TrainEpoch}{$\mathcal{T}^0, {\Theta^0}$}
    \State $\Theta\leftarrow\Theta^0$
    \For {$q\leftarrow\mathcal{N}^0-{\rm root}\left(\mathcal{T}^0\right)$} \Comment{Root is not used as query}
        \State $\mathcal{X}_q=\{\}$ \Comment{Initialize anchor set}
        \State $p\leftarrow{\rm parent}\left(q\right)$ \Comment{Reference node of labeling}
        \State $\mathcal{X}_q\leftarrow\mathcal{X}_q\cup\left\{p\right\}$ \Comment{Ground Truth Parent: $S_p=1, S_c=1$}
        \State $\mathcal{X}_q\leftarrow\mathcal{X}_q\cup{\rm sample}\left(\mathcal{A}_p\right)$ \Comment{Ancestors: $S_p=1, S_f=1$}
        \State $\mathcal{X}_q\leftarrow\mathcal{X}_q\cup{\rm sample}\left(\mathcal{D}_p\right)$ \Comment{Descendants: $S_p=1, S_b=1$}
        \State $\mathcal{X}_q\leftarrow\mathcal{X}_q\cup{\rm sample}\left(\mathcal{N}^0-\left\{p\right\}-\mathcal{A}_p-\mathcal{D}_p\right)$\\
        \Comment{Other nodes: $S_p=0, S_b=1$}
        \For {$a\leftarrow\mathcal{X}_q$}
            \State Compute $S_p\left(a,q\right)$ using Eqn. \ref{eqn:pathfinder}
            \State Compute $S_f\left(a,q\right),S_c\left(a,q\right),S_b\left(a,q\right)$ using Eqn. \ref{eqn:stopper}
        \EndFor
        \State Compute $\mathcal{L}_q$ with $S_p,S_f,S_c,S_b$ using Eqn. \ref{eqn:loss}
        \State $\Theta\leftarrow{\rm optimize}\left(\Theta,\mathcal{L}_q\right)$
    \EndFor
    \State \textbf{return} $\Theta$
\EndProcedure
\end{algorithmic}
\end{algorithm}

% \begin{eqnarray}
%     \mathcal{L}_q=-\frac{1}{\mathopen|\mathcal{X}\mathclose|}\left(\sum_{a\in\mathcal{X}}{\hat{S_p}\left(a,q\right)\log S_p\left(a,q\right)+\left(1-\hat{S_p}\left(a,q\right)\right)\log \left(1-S_p\left(a,q\right)\right)+\eta\sum_{k\in\left\{f,c,b\right\}}{\hat{s_k}\left(a,q\right)\log s_k\left(a,q\right)}}\right)
% \end{eqnarray}

\textbf{Fitting Score-based Inference.} During inference, evaluation of an anchor-query pair $\left<a,q\right>$ should consider both Pathfinder's path evaluation and Stopper's level evaluation. However, instead of merely using $S_p$ and $S_c$, the multi-classifying Stopper also enables the \textsf{HEF} model to disambiguate the most suited anchor from its neighbors (its direct parent and children) and self-correct its level prediction by exchanging scores with its neighbors to find the best position for maintaining the taxonomy's coherence. Thus, We introduce the {\em Fitting Score} function during inference. For a new query term $q\in\mathcal{C}$, we first obtain the Pathfinder Scores and Stopper Scores of all node pairs $\left<a,q\right>, a\in \mathcal{N}^0$. For each anchor node $a$, we assign its Fitting Score by multiplying the following four items:
\begin{itemize}
    \item \textbf{$a$'s Pathfinder Score: $S_p\left(a,q\right)$}, which suggests whether $a$ is on the right path.
    \item \textbf{$a$'s parent's Forward Score: $S_f\left({\rm parent}\left(a\right),q\right)$}, which distinguishes $a$ and $a$'s parent, and rectifies $a$'s Current Score. When $a$ is the root node, we assign this item as a small number like $1e-4$ since the first level of taxonomy is likely to remain unchanged.
    \item \textbf{$a$'s Current Score: $S_c\left(a,q\right)$}, which suggests whether $a$ is on the right level.
    \item \textbf{$a$'s child with maximum Pathfinder Score's Backward Score: $S_b\left(c_a^*,q\right), c_a^*=\mathop{\arg\max}_{c_a\in{\rm child}\left(a\right)}{S_p\left(c_a,q\right)}$}, which distinguishes $a$ and $a$'s children, and rectifies $a$'s Current Score. Since $a$ might have multiple children, we pick the child with max Pathfinder Score, for larger $S_p$ indicates a better hypernymy relation to $q$. When $a$ is a leaf node, we assign this item as the proportion of leaf nodes in the seed taxonomy to keep its overall design.
\end{itemize}
The Fitting Score of $\left<a,q\right>$ is given by:
\begin{equation}
    F\left(a,q\right)=S_p\left(a,q\right)\cdot S_f\left({\rm parent}\left(a\right),q\right)\cdot S_c\left(a,q\right)\cdot S_b\left(c_a^*,q\right)
    \label{eqn:fittingscore}
\end{equation}
\begin{displaymath}
    c_a^*=\mathop{\arg\max}_{c_a\in{\rm child}\left(a\right)}{S_p\left(c_a,q\right)}.
\end{displaymath}

The Fitting Score can be computed using ordered $S_p,S_f,S_c,S_b$ arrays and the seed taxonomy's adjacency matrix. Since a tree's adjacency matrix is sparse, the time complexity of Fitting Score computation is low. After calculating the Fitting Scores between all seed nodes and the query, we select the seed node with the highest Fitting Score as the query's parent in the expanded taxonomy:
\begin{equation}
    {\rm parent}\left(q\right)\coloneqq\mathop{\arg\max}_{a\in\mathcal{N}^0}F\left(a,q\right).
\end{equation}

\section{Experiments}

In this section, we first introduce our experimental setups, including datasets, our implementation details, evaluation criteria, and a brief description of the compared baseline methods. Then, we provide extensive evaluation results for overall model performance, performance contribution brought by each design, and sensitivity analysis of the multi-task learning weight $\eta$ in Equation~\ref{eqn:loss}. In-depth visualizations of hypernymy detection and coherence modeling modules are provided to analyze the model's inner behavior. We also provide a case study in the appendix.\footnote{The code will be available at \url{https://github.com/sheryc/HEF}.}

\subsection{Experimental Setup}

\subsubsection{Datasets}

We evaluate \textsf{HEF} on three public benchmark datasets retrieved from SemEval-2016 Task 13 \cite{bordea_semeval-2016_2016}. This task contains three taxonomies in the domain of Environment ({\em SemEval16-Env}), Science ({\em SemEval16-Sci}), and Food ({\em SemEval16-Food}), respectively. The statistics of the benchmark datasets are provided in Table \ref{tab:dataset}. Note that the original dataset may not form a tree. In this case, we use a spanning tree of the taxonomy instead of the original graph to match the problem definition. The pruning process only removes less than 6\% of the total edges, keeping the taxonomy's information and avoiding multiple ground truth parents for a single node.

Since \textsf{HEF} and the compared baselines \cite{shen_taxoexpan_2020,yu_steam_2020} are all limited to adding new terms without modifying the seed taxonomy, nodes in the test and validation set can only sample from leaf nodes to guarantee that the parents of test or validation nodes exist in the seed taxonomy. This is also the sampling strategy of TaxoExpan \cite{shen_taxoexpan_2020}. Following the previous state-of-the-art model STEAM \cite{yu_steam_2020}, we exclude 20\% of the nodes in each dataset, of which ten nodes of each dataset are separated as the validation set for early stopping, and the rest as the test set. The nodes not included in the validation set and test set are seed nodes for self-supervision in the training phase and potential anchor nodes in the inference phase. Note that pruning the dataset does not affect the node count, thus the scale of the dataset remains identical to our baselines' settings.

\begin{table}[t]
  \caption{Statistics of datasets. $\left|N\right|$ and $\left|E_O\right|$ are the numbers of nodes and edges in the original datasets, respectively. $D$ is the depth of the taxonomy. We adopt the spanning tree of each dataset, and $\left|E\right|$ is the number of remaining edges.}
  \label{tab:dataset}
  \begin{tabular}{ccccc}
  \toprule
  Dataset & $\left|N\right|$ & $\left|E_O\right|$ & $\left|E\right|$ & $D$\\
    \midrule
    {\em SemEval16-Env} & 261 & 261 & 260 & 6\\
    {\em SemEval16-Sci} & 429 & 452 & 428 & 8\\
    {\em SemEval16-Food} & 1486 & 1576 & 1485 & 8\\
  \bottomrule
\end{tabular}
\end{table}

\begin{table*}[t]
    \caption{Comparison of the proposed method against state-of-the-art methods. All metrics are presented in percentages (\%). Best results for each metric of each dataset are marked in \textbf{bold}. Reported performance is the average of three runs using different random seeds. The MRR of TAXI \cite{panchenko_taxi_2016} is inaccessible since it outputs the whole taxonomy instead of node rankings. The performance of baseline methods are retrieved from \cite{yu_steam_2020}.}
    \label{tab:main}
    \begin{tabular}{c|ccc|ccc|ccc}
        \toprule
        Dataset & \multicolumn{3}{c|}{\em SemEval16-Env} & \multicolumn{3}{c|}{\em SemEval16-Sci} & \multicolumn{3}{c}{\em SemEval16-Food}\\
        Metric & Acc & MRR & Wu\&P & Acc & MRR & Wu\&P & Acc & MRR & Wu\&P \\
        \midrule
        BERT+MLP & 11.1 & 21.5 & 47.9 & 11.5 & 15.7 & 43.6 & 10.5 & 14.9 & 47.0 \\
        TAXI \cite{panchenko_taxi_2016} & 16.7 & - & 44.7 & 13.0 & - & 32.9 & 18.2 & - & 39.2 \\
        HypeNet \cite{shwartz_improving_2016} & 16.7 & 23.7 & 55.8 & 15.4 & 22.6 & 50.7 & 20.5 & 27.3 & 63.2 \\
        TaxoExpan \cite{shen_taxoexpan_2020} & 11.1 & 32.3 & 54.8 & 27.8 & 44.8 & 57.6 & 27.6 & 40.5 & 54.2 \\
        STEAM \cite{yu_steam_2020} & 36.1 & 46.9 & 69.6 & 36.5 & 48.3 & 68.2 & 34.2 & 43.4 & 67.0 \\
    \midrule
    \textsf{HEF} & \textbf{55.3} & \textbf{65.3} & \textbf{71.4} & \textbf{53.6} & \textbf{62.7} & \textbf{75.6} & \textbf{47.9} & \textbf{55.5} & \textbf{73.5} \\
    \bottomrule
  \end{tabular}
\end{table*}

\subsubsection{Baselines for Comparison}

We compare our proposed \textsf{HEF} model with the following baseline approaches:

\begin{itemize}
    \item \textbf{BERT+MLP}: This method utilizes BERT \cite{devlin_bert_2019} to perform hypernym detection. This model's input is the term's surface name, and the representation of BERT's classification token {\em $\langle$CLS$\rangle$} is fed into a feed-forward layer to score whether the first sequence is the ground-truth parent.
    \item \textbf{HypeNet} \cite{shwartz_improving_2016}: HypeNet is an LSTM-based hypernym extraction model that scores a term pair by representing node paths in the dependency tree.
    \item \textbf{TAXI} \cite{panchenko_taxi_2016}: TAXI was the top solution of SemEval-2016 Task 13. It explicitly splits the task into a pipeline of hypernym detection using substring matching and pattern extraction, and hypernym pruning to avoid multiple parents.
    \item \textbf{TaxoExpan} \cite{shen_taxoexpan_2020}: TaxoExpan is a self-supervised taxonomy expansion model. The anchor's representation is modeled by a graph network of its Egonet with consideration of relative levels, and the parental relationship is scored by a feed-forward layer. BERT embedding is used as its input instead of the model's original configuration.
    \item \textbf{STEAM} \cite{yu_steam_2020}: STEAM is the state-of-the-art self-supervised taxonomy expansion model, which scores parental relations by ensembling three classifiers considering graph, contextual, and hand-crafted lexical-syntactic features, respectively.
\end{itemize}

\subsubsection{Implementation Details}

In our setting, the coherence modeling module is a 3-layer, 6-head, 768-dimensional Transformer encoder initialized from Gaussian distribution $\mathcal{N}\left(0, 0.02\right)$. The first hidden layers of Pathfinder and Stopper are both 300-dimensional. The input to the hypernymy detection module is either truncated or padded to a length of 64. Each training step contains a set of 32 query-ego-tree pairs of 32 query nodes using gradient accumulation, with each query-ego-tree pair set containing one ground-truth parent ($S_p=1$, $S_c=1$), at most 6 ground-truth parent's ancestors ($S_p=1$, $S_f=1$), at most 8 ground-truth parent's descendants ($S_p=1$, $S_b=1$), and at least 16 other nodes ($S_p=0$, $S_b=1$). The hyperparameters above are empirically set, since our algorithm is not sensitive to the setting of splits. Each dataset is trained for 150 epochs. In a single epoch, each seed node is trained as the query exactly once. AdamW \cite{loshchilov_decoupled_2019} is used for optimization with $\epsilon$ set to $1\times 10^{-6}$. A linear warm-up is adopted with the learning rate linearly rise from 0 to 5e-5 in the first 10\% of total training steps and linearly drop to 0 at the end of 150 epochs. The multi-task learning weight $\eta$ is set to 0.9. After each epoch, we validate the model and save the model with the best performance on the validation set. These hyperparameters are tuned on on {\em SemEval16-Env}'s validation set, and are used across all datasets and experiments unless specified in the ablation studies or sensitivity analysis.

\subsubsection{Evaluation Metrics}

Assume $k\coloneqq\mathopen|\mathcal{C}\mathclose|$ to be the term count of the test set, $\left\{p_1, p_2, \cdots, p_k\right\}$ to be the predicted parents for test set queries, and $\left\{\hat{p_1}, \hat{p_2}, \cdots, \hat{p_k}\right\}$ to be the ground truth parents accordingly. Following the previous solutions \cite{manzoor_expanding_2020, vedula_enriching_2018, yu_steam_2020}, we adopt the following three metrics as evaluation criteria.

\begin{itemize}
    \item \textbf{Accuracy (Acc)}: It measures the proportion that the predicted parent for each node in the test set exactly matches the ground truth parent:
    \begin{displaymath}
        \mbox{Acc}=\mbox{Hit@1}=\frac{1}{k}\sum_{i=1}^{k}{\mathbb{I}\left(p_i=\hat{p_i}\right)},
    \end{displaymath}
    \noindent where $\mathbb{I}(\cdot)$ denotes the indicator function,
    \item \textbf{Mean Reciprocal Rank (MRR)}: It calculates the average reciprocal rank of each test node's ground truth parent:
    \begin{displaymath}
        \mbox{MRR}=\frac{1}{k}\sum_{i=1}^{k}{\frac{1}{\mbox{rank}\left(\hat{p_i}\right)}},
    \end{displaymath}
    \item \textbf{Wu \& Palmer Similarity  (Wu\&P) }\cite{wu_verbs_1994}: It is a tree-based measurement that judges how close the predicted and ground truth parents are in the seed taxonomy:
    \begin{displaymath}
        \mbox{Wu\&P}=\frac{1}{k}\sum_{i=1}^{k}{\frac{2\times \mbox{depth}\left(\mbox{LCA}\left(p_i, \hat{p_i}\right)\right)}{\mbox{depth}\left(p_i\right)+\mbox{depth}\left(\hat{p_i}\right)}},
    \end{displaymath}
    \noindent where $\mbox{depth}(\cdot)$ is the node's depth in the seed taxonomy and $\mbox{LCA}(\cdot,\cdot)$ is the least common ancestor of two nodes.
\end{itemize}

\subsection{Main Results}

The performance of \textsf{HEF} is shown in Table \ref{tab:main}. \textsf{HEF} achieves the best performance on all datasets and surpasses previous state-of-the-art models with a significant improvement on all metrics.

From the table, we get an overview of how taxonomy expansion models evolve chronologically. The solution of BERT+MLP does not utilize any structural and lexical-syntactic features of terms, and the insufficiency of information attributes to its poor results. Models of the first generation like TAXI and HypeNet utilize lexical, syntactic, or contextual information to achieve better results, mainly for the task's hypernymy detection part. However, these two models do not utilize any of the structural information of taxonomy; hence they are unable to maintain the taxonomy's structural design. Models of the second generation, like TaxoExpan and STEAM, inherit handcrafted lexical-syntactic features for detecting hypernymy relations. They also utilize the structural information by self-supervision from seed taxonomy and graph neural networks on small subgraphs of the taxonomy. However, they neglect the hierarchical structure of taxonomies, and they do not consider the coherence of the whole expanded taxonomy. Thus, their usage of structural information is only an improvement for performing hypernymy detection rather than taxonomy expansion.

\textsf{HEF} further improves both the previous two generations' strength by proposing a new approach that better fits the taxonomy expansion task. Moreover, it introduces a new goal for the task: to best preserve the taxonomy's coherence after expansion. We propose the description generation algorithm to generate accurate and domain-specific descriptions for complex and rare terms, to incorporate lexical-syntactic features for hypernymy detection. Aided by DistilBERT's power of sentence-pair representation, \textsf{HEF} can mine hypernymy features more automatically and accurately. \textsf{HEF} also aims to fully exploit the information brought by the taxonomy's hierarchical structure to boost performance. \textsf{HEF} uses ego-trees to perform thorough comparison along root path and among siblings, injects tree-exclusive features to assist modeling the expert-curated taxonomy designs and explicitly evaluates both path and level for the anchor node as well as its parent and child. Experiment results suggest that these designs are capable of modeling and maximizing the coherence of taxonomy in different aspects, which results in a vast performance increase in the taxonomy expansion task.

% Solving all the problems described above, \textsf{HEF} utilizes the description of terms as input and utilize DistilBERT's power of sentence-pair relation to mine hypernymy relation in a more direct fashion. The coherence modeling module's ability of integrating tree-exclusive features and information of all nodes related to the anchor also contributes to more in-depth understanding of taxonomy structure alignment. Lastly, the fitting score calculation explicitly introduces a node's parent and child's information for direct comparison, as well as integrating both path and level for accurate vertical and horizontal selection, which largely enhances the model's ranking ability for the taxonomy expansion task.

\begin{figure*}[t]
\begin{tabular}{p{0.655\columnwidth}p{0.655\columnwidth}p{0.655\columnwidth}}
    \includegraphics[clip, trim=0.41cm 0cm 1.52cm 1.2cm, width=0.655\columnwidth]{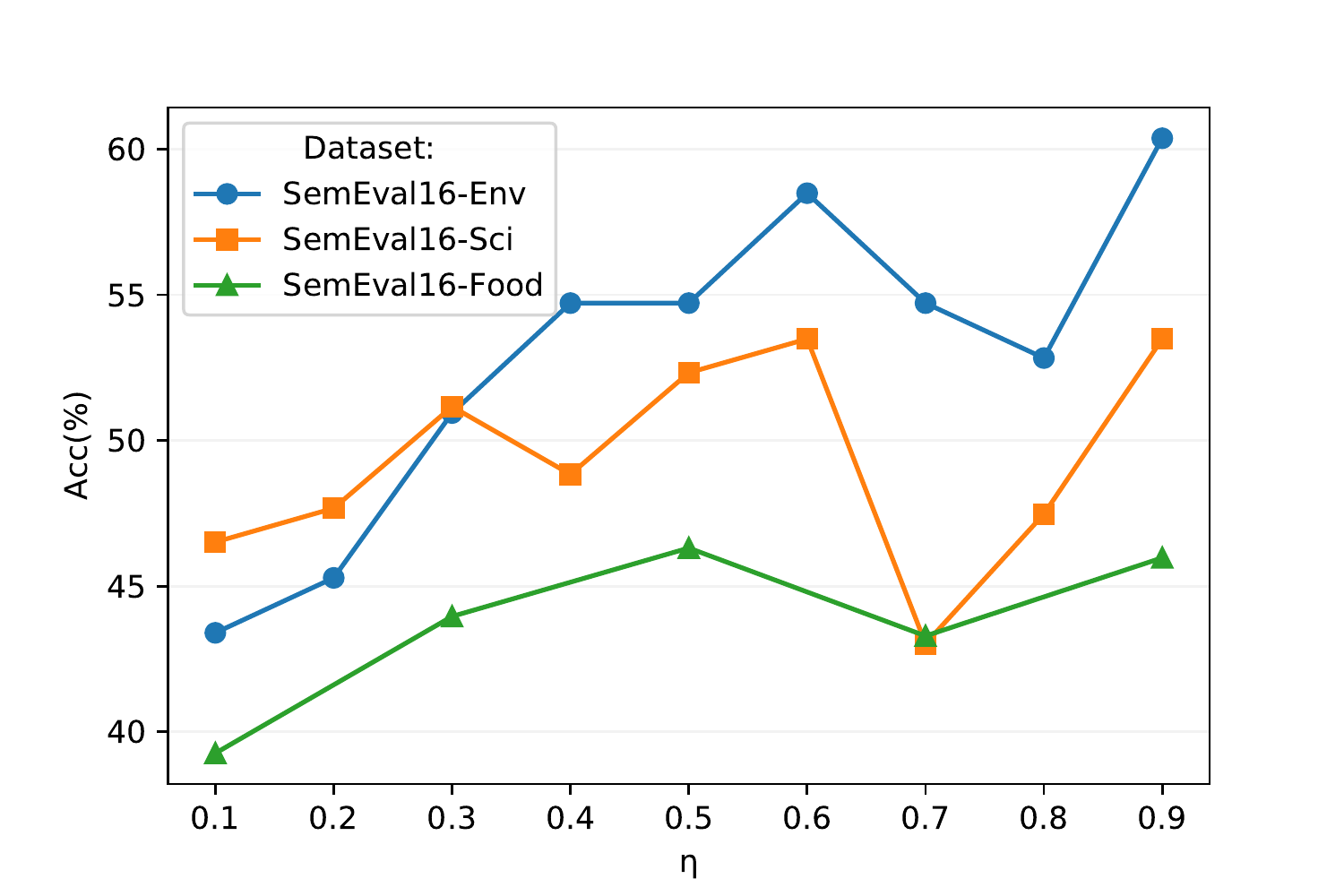} & 
    \includegraphics[clip, trim=0.41cm 0cm 1.52cm 1.2cm, width=0.655\columnwidth]{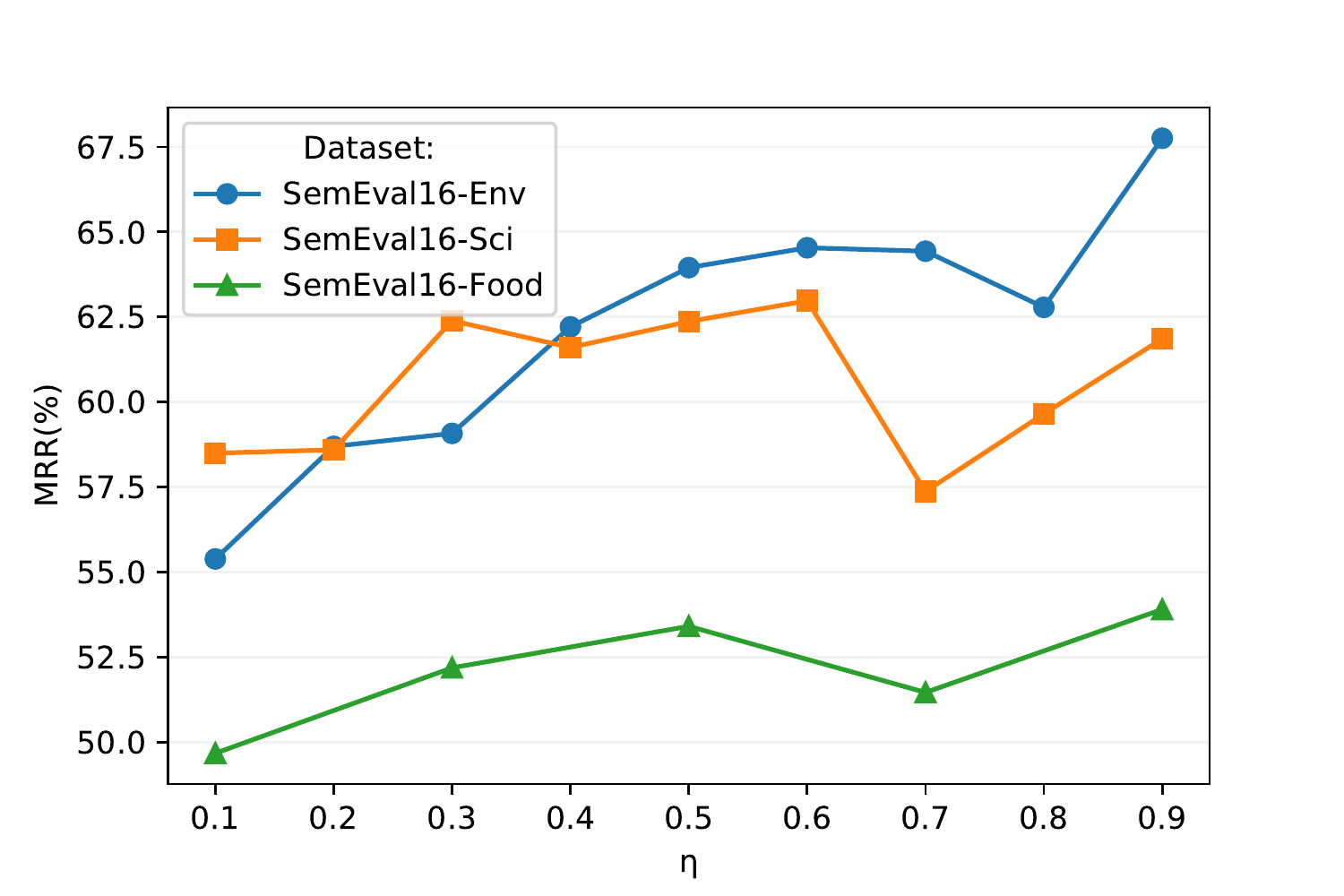} & 
    \includegraphics[clip, trim=0.41cm 0cm 1.52cm 1.2cm, width=0.655\columnwidth]{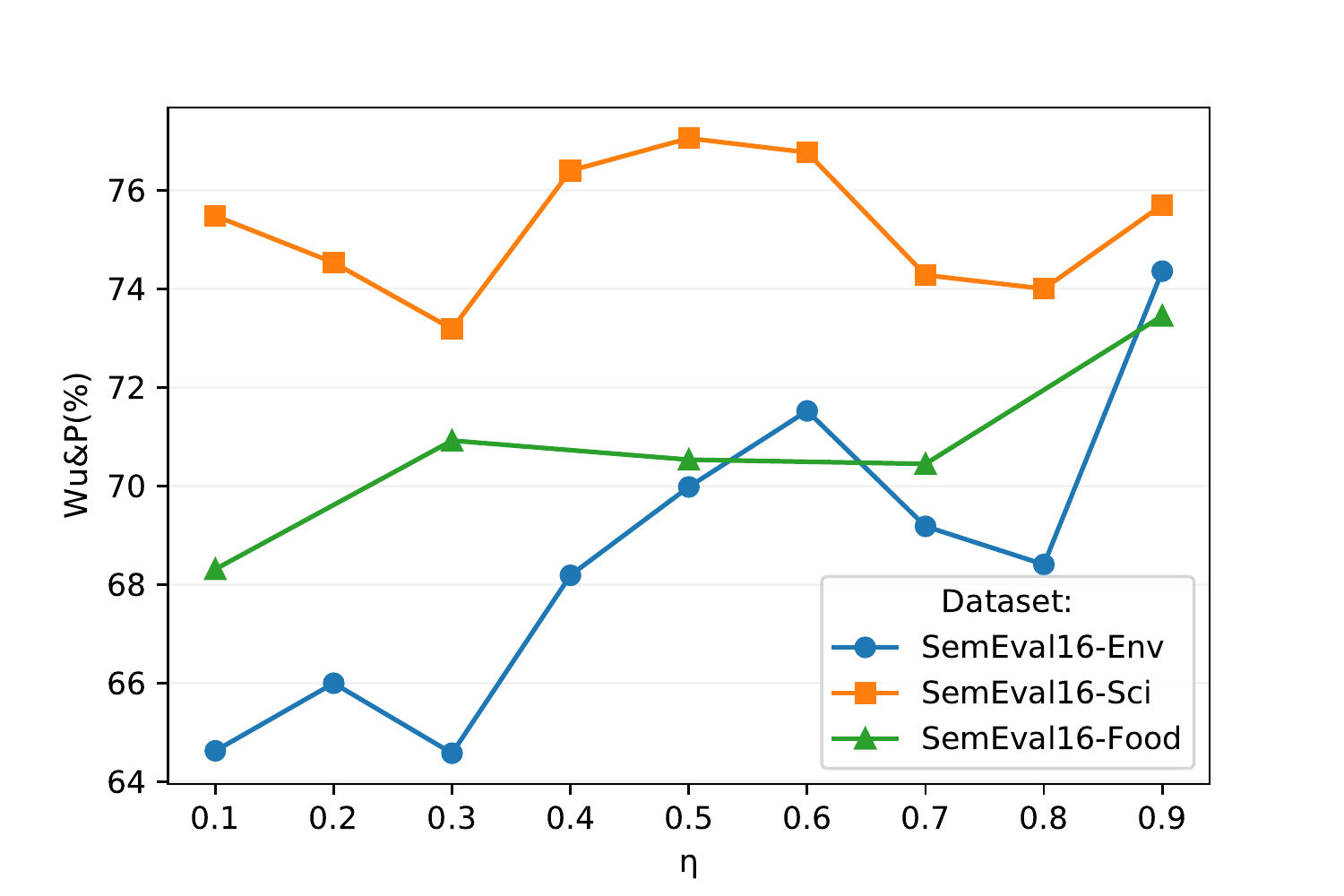} \\
    \makecell[c]{(a) Accuracy on all 3 datasets.} & 
    \makecell[c]{(b) MRR on all 3 datasets.} & 
    \makecell[c]{(c) Wu\&P on all 3 datasets.} \\
\end{tabular}
\caption{Sensitivity analysis of model performance under different multi-task learning weight $\eta$.}
\label{fig:eta}
\end{figure*}

\subsection{Ablation Studies}

\begin{table}[t]
    \caption{Ablation experiment results on the {\em SemEval16-Env} dataset. All metrics are presented in percentages (\%). For each ablation experiment setting, only the best result is reported.}
    \label{tab:ablation}
    \begin{tabular}{c|c|ccc}
        \toprule
        Abl. Type & Setting & Acc & MRR & Wu\&P \\
        \midrule
        Original & \makecell[c]{\textsf{HEF}} & 55.3 & 65.3 & 71.4 \\
        \midrule
        \multirow{4}{*}{Dataflows} & \makecell[l]{- WordNet Descriptions} & 41.5 & 55.3 & 62.6 \\
        & \makecell[l]{- Ego-tree + Egonet} & 45.3 & 60.6 & 69.9 \\
        & \makecell[l]{- Relative Level Emb.} & 49.1 & 59.2 & 60.9 \\
        & \makecell[l]{- Absolute Level Emb.} & 49.1 & 60.6 & 68.4 \\
        \midrule
        \multirow{3}{*}{\makecell{Scoring\\ Function}} & \makecell[l]{Stopper Only} & 52.8 & 62.5 & 68.7 \\
        & \makecell[l]{Pathfinder + Current Only} & 50.9 & 62.1 & 66.8\\
        & \makecell[l]{Current Only} & 41.5 & 54.7 & 58.6\\
        \bottomrule
    \end{tabular}
\end{table}

We discuss how exploiting different characteristics of taxonomy's hierarchical structure brings performance increase by a series of ablation studies. We substitute some designs of \textsf{HEF} in dataflow and score function to a vanilla setting and rerun the experiments. The results of the ablation studies are shown in Table~\ref{tab:ablation}.

% To validate the contribution of various modules and design towards \textsf{HEF}'s performance, we conduct ablation experiments for the pretrained hypernymy detection module, several designs of dataflow and our Fitting Score calculation.

\begin{itemize}
    \item \textbf{- WordNet Descriptions}: WordNet descriptions are substituted with the term's surface name as the hypernymy detection module's input.
    \item \textbf{- Ego-tree + Egonet}: the Egonet from TaxoExpan \cite{shen_taxoexpan_2020} is used instead of the ego-tree for modeling the tree structure.
    \item \textbf{- Relative Level Emb.}: The relative level embedding for the coherence modeling module is removed.
    \item \textbf{- Absolute Level Emb.}: The absolute level embedding for the coherence modeling module is removed.
    \item \textbf{Stopper Only}: Only the Stopper Scores are used for Fitting Score calculation. More specifically, $\eta=0$, and the Fitting Score in Equation \ref{eqn:fittingscore} becomes:
    \begin{displaymath}
        F\left(a,q\right)=S_f\left({\rm parent}\left(a\right),q\right)\cdot S_c\left(a,q\right)\cdot S_b\left(c_a^*,q\right),
    \end{displaymath}
    \begin{displaymath}
        c_a^*=\mathop{\arg\max}_{c_a\in{\rm child}\left(a\right)}{S_p\left(c_a,q\right)}.
    \end{displaymath}
    \item \textbf{Pathfinder + Current Only}: Only the Pathfinder Score and the Current Score are used for Fitting Score calculation. More specifically, the Fitting Score in Equation \ref{eqn:fittingscore} and the loss in Equation \ref{eqn:loss} become:
    \begin{displaymath}
        F\left(a,q\right)=S_p\left(a,q\right)\cdot S_c\left(a,q\right),
    \end{displaymath}
    \begin{align*}
        \mathcal{L}_q=&-\eta\frac{1}{\mathopen|\mathcal{X}_q\mathclose|}\sum_{a\in\mathcal{X}_q}{{\rm BCELoss}\left(\hat{S_p}\left(a,q\right), S_p\left(a,q\right)\right)}\\[1ex]
        &-\left(1-\eta\right)\frac{1}{\mathopen|\mathcal{X}_q\mathclose|}\sum_{a\in\mathcal{X}_q}{{\rm BCELoss}\left(\hat{S_c}\left(a,q\right), S_c\left(a,q\right)\right)}.
    \end{align*}
    \item \textbf{Current Only}: Only the Current Score is used for Fitting Score calculation. This is the scoring strategy identical to prior arts \cite{shen_taxoexpan_2020, yu_steam_2020}. More specifically, the Fitting Score in Equation \ref{eqn:fittingscore} and the loss in Equation \ref{eqn:loss} become:
    \begin{displaymath}
        F\left(a,q\right)=S_c\left(a,q\right),
    \end{displaymath}
    \begin{displaymath}
        \mathcal{L}_q=-\frac{1}{\mathopen|\mathcal{X}_q\mathclose|}\sum_{a\in\mathcal{X}_q}{{\rm BCELoss}\left(\hat{S_c}\left(a,q\right), S_c\left(a,q\right)\right)}.
    \end{displaymath}
\end{itemize}

We notice that by changing the design of dataflows, the performance of the \textsf{HEF} model suffers from various deteriorations. Substituting WordNet descriptions with a term's surface name surprisingly remains a relatively high performance, which might attribute to the representation power of the DistilBERT model. Using Egonets rather than ego-trees for coherence modeling also affects the performance. Although egonets can capture the local structure of taxonomy, ego-trees are more capable of modeling the complete construction of a hierarchy. For the introduction of level embeddings, the results show that removing one of the two level embeddings for the coherence modeling module hurts the learning of taxonomy's design. This is in accordance with the previous research about the importance of using the information of both absolute and relative positions in Transformers \cite{shaw_self-attention_2018} and confirms our assumption that taxonomies have intrinsic designs about both absolute and relative levels.

Changes to the score function bring a smaller negative impact on the model compared to the dataflow changes, except for the setting of using merely Current Score. When using only the Current Score, the model loses the ability to disambiguate with its neighbors and the capacity of directly choosing the right path, downgrading the problem to be a series of independent scoring problems like the previous solutions. Adding Backward Score and Forward Score into Fitting Score calculation allows the model to distinguish the ground truth from its neighbors, bringing a boost to accuracy. However, without the Pathfinder, the ``Stopper Only'' setting only explicitly focuses on choosing the right level without considering the path and is inferior to the original \textsf{HEF} model.

However, we observe that although changing several designs of dataflow or scoring function deteriorates the performance, our method can still surpass the previous state-of-the-art in Acc and MRR, suggesting that the \textsf{HEF} model introduces improvements in multiple aspects, which also testifies that maximizing the taxonomy's coherence is a better goal for the taxonomy expansion task.

\begin{figure}[t]
  \centering
  \includegraphics[width=0.83\linewidth]{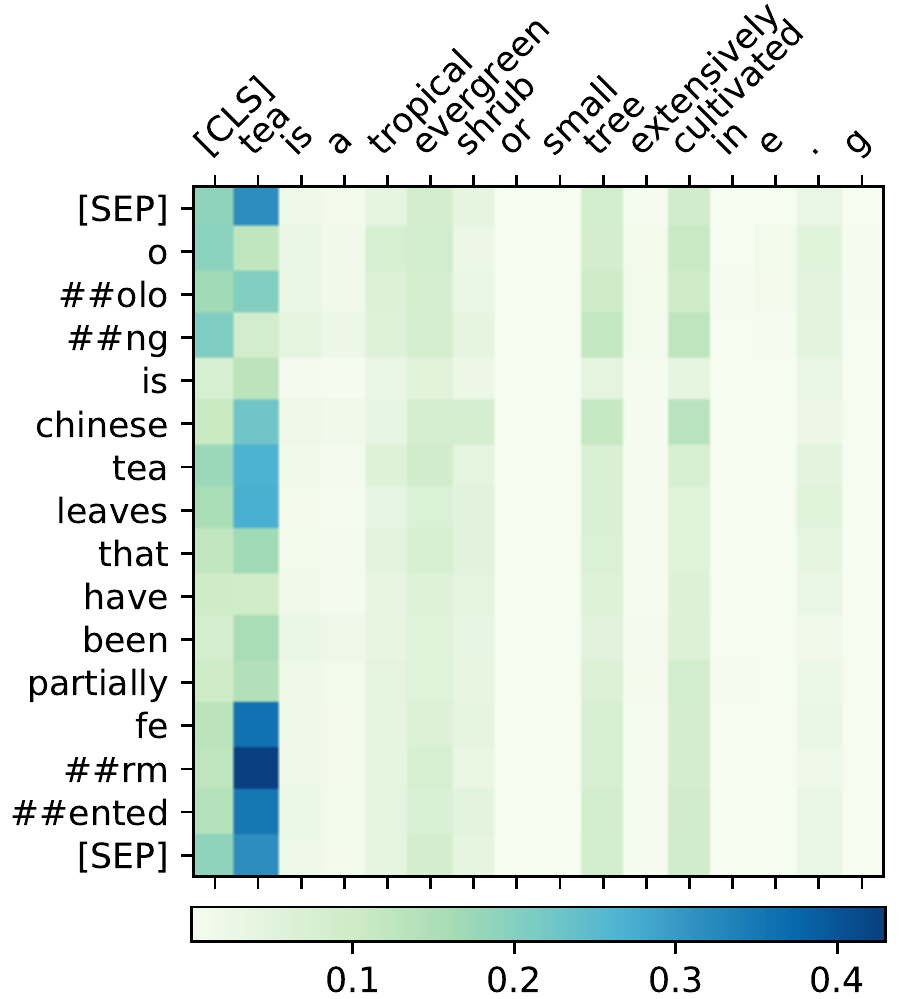}
  \caption{Illustration of one self-attention head in the last layer of the hypernymy detection module, showing how the hypernymy detection module detects hypernymy relations. In this example, the seed node is {\em ``tea''}, and the query is {\em ``oolong''}.}
  \label{fig:lower}
\end{figure}

\subsection{Impact of Multi-Task Learning Weight $\eta$}

In this section, we discuss the impact of $\eta$ in Equation \ref{eqn:loss} through a sensitivity analysis. Since $\eta$ controls the proportion of loss from the path-oriented Pathfinder and the level-oriented Stopper, this hyperparameter affects \textsf{HEF}'s tendency to prioritize path or level selection. The results on all three datasets are shown in Fig.~\ref{fig:eta}.

From the result, we notice that $\eta$ cannot be set too low, which means that explicit path selection contributes a lot to the model's performance. This is in accordance with the fact that taxonomies are based on hypernymy relations and selecting the right path is the essential guidance for anchor selection. A better setting of $\eta$ is $\left[0.4, 0.6\right]$. In this setting, the model tends to balance path and level selections, which results in better performance. Surprisingly, setting $\eta$ to a high value like 0.9 also brings a performance boost, and sometimes even achieves the best result. This phenomenon consistently exists when changing random seeds. However, setting $\eta$ to 1 means using merely Pathfinder, which cannot distinguish the ground truth from other nodes and breaks the model. This discovery further testifies the importance of explicitly evaluating path selection in the taxonomy expansion task.

\subsection{Visualization of Self-Attentions}

\subsubsection{Node Pair Hypernymy Detection Module}

To illustrate how the hypernymy detection module works, we show the weights of one of the attention heads of the hypernymy detection module's last Transformer encoder layer in Fig.~\ref{fig:lower}.

By expanding a term to its description, the model is capable of understanding the term {\em ``oolong''} by its description, which cannot be achieved by constructing rule-based lexical-syntactic features since {\em ``oolong''} and {\em ``tea''} have no similarity in their surface names. Furthermore, by adopting the pretrained DistilBERT, the hypernymy detection module can also discover more latent patterns such as the relation between ``leaves'' and ``tree'', allowing the model to discover more in-depth hypernymy relations.

% By expanding a term to its description, the model is capable of understanding the uncommon term "oolong" by its description, which cannot be achieved by constructing rule-based lexical-syntactic features. By adopting pretrained DistilBERT, the model can also discover the relation between ``leaves'' and ``tree'', allowing the model to discover more latent hypernymy patterns. Lorem ipsum dolor sit amet, consectetur adipiscing elit, sed do eiusmod tempor incididunt ut labore et dolore magna aliqua.

\subsubsection{Ego-Tree Coherence Modeling Module}

\begin{figure}[t]
  \centering
  \includegraphics[width=0.9\linewidth]{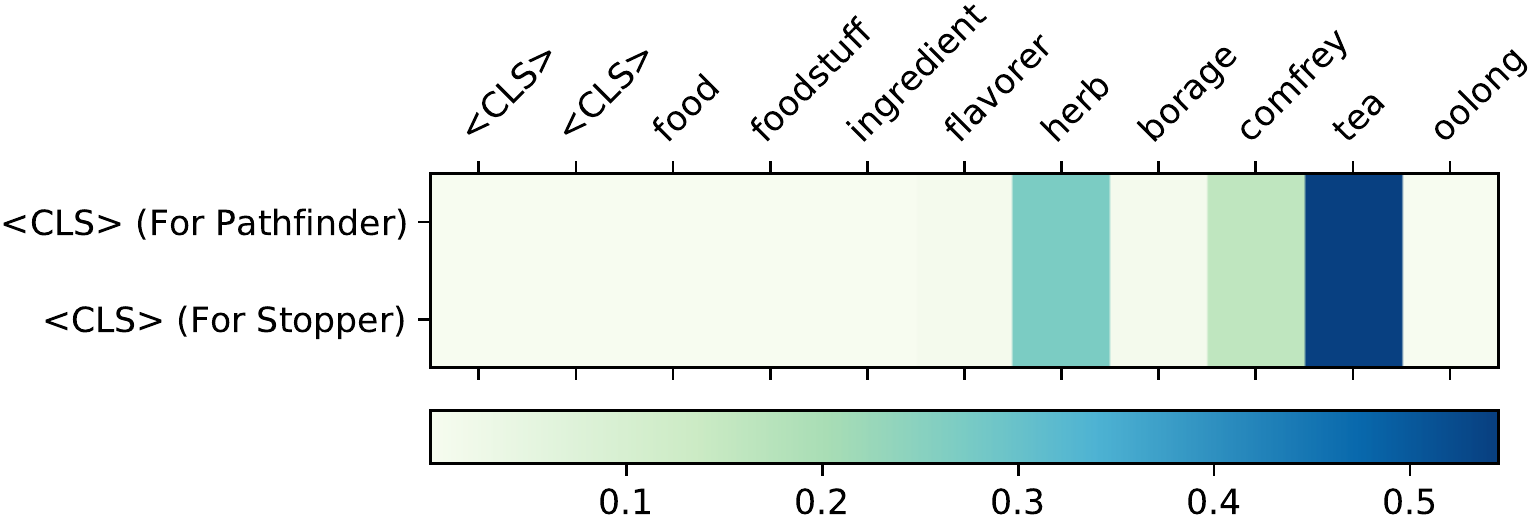}
  \caption{Illustration of one self-attention head in the first layer of the coherence modeling module, showing how the coherence modeling module finds the most fitted node in an ego-tree. In this example, the anchor is {\em ``herb''}, the query is {\em ``oolong''}, and the query's ground truth parent is {\em ``tea''}.}
  \label{fig:higher}
\end{figure}

To illustrate how the coherence modeling module compares the nodes in the ego-tree to maintain the taxonomy's coherence, we present the weights of an attention head of the module's first Transformer encoder layer in Fig.~\ref{fig:higher}. Since the last layer's attention mostly focuses on the anchor node ({\em ``herb''}), the first layer can better illustrate the model's comparison among ego-tree nodes.

Based on our observation, the two {\em <CLS>}s are capable of finding the best-suited parent node in the ego-tree even if it is not the anchor. Since the coherence modeling module utilizes ego-trees for hierarchy modeling, the coherence modeling module can compare a node with all its ancestors and its children to find the most suited anchor, which makes the model more robust. Besides, the coherence modeling module is also able to assign a lower attention weight to the best-suited parent's parent node when it is on the right path, suggesting that the coherence modeling module can achieve both path-wise and level-wise comparison.

\section{Conclusion}

% We proposed \textsf{HEF}, a self-supervised taxonomy expansion model that fully exploits the hierarchical structure of a taxonomy for better hierarchical structure modeling and taxonomy coherence maintenance. It learns the design of taxonomy by using hierarchical ego-trees for self-supervision, models the coherence of taxonomy by integrating tree-exclusive features into a tree-modeling transformer encoder, and evaluates the coherence of taxonomy by Fitting Score based on explicit path and level evaluation as well as the selection of anchor's parent and child's level. Extensive experiments suggest that by fully exploiting the hierarchical structure of taxonomies for self-supervision, feature injection, multi-dimensional evaluation, disambiguation and coherence maintenance, \textsf{HEF} is able to generate more coherent taxonomy than previous solutions and consistently outperforms the previous state-of-the-art on three benchmark taxonomy expansion datasets in different domains by a large margin.

% Below is the version proposed in the rebuttal

We proposed \textsf{HEF}, a self-supervised taxonomy expansion model that fully exploits the hierarchical structure of a taxonomy for better hierarchical structure modeling and taxonomy coherence maintenance. Compared to previous methods that evaluate the anchor by merely a new edge in a normal graph neglecting the tree structure of taxonomy, we used extensive experiments to prove that, evaluating a tree structure for coherence maintenance, and mining multiple tree-exclusive features in the taxonomy, including hypernymy relations from parent-child relations, term similarity from sibling relations, absolute and relative levels, path+level based multi-dimensional evaluation, and disambiguation based on parent-current-child chains all brought performance boost. This indicates the importance of using the information of tree structure for the taxonomy expansion task. We also proposed a framework for injecting these features, introduced our implementation of the framework, and surpassed the previous state-of-the-art. We believe that these novel designs and their motivations will not only benefit the taxonomy expansion task, but also be influential for all tasks involving hierarchical or tree structure modeling and evaluation. Future works include how to model and utilize these or new tree-exclusive features to boost other taxonomy-related tasks, and better implementation of each module in HEF.

% Lorem ipsum dolor sit amet, consectetur adipiscing elit, sed do eiusmod tempor incididunt ut labore et dolore magna aliqua. Ut enim ad minim veniam, quis nostrud exercitation ullamco laboris nisi ut aliquip ex ea commodo consequat. Duis aute irure dolor in reprehenderit in voluptate velit esse cillum dolore eu fugiat nulla pariatur. Excepteur sint occaecat cupidatat non proident, sunt in culpa qui officia deserunt mollit anim id est laborum. Lorem ipsum dolor sit amet, consectetur adipiscing elit, sed do eiusmod tempor incididunt ut labore et dolore magna aliqua. 

\begin{acks}
Thanks to everyone who helped me with this paper in the Tencent Jarvis Lab, my family, and my loved one.
\end{acks}

\bibliographystyle{ACM-Reference-Format}
\bibliography{references}

\appendix

\section{Case Study}

To understand how different Fitting Score components contribute to \textsf{HEF}'s performance, we conduct a case study on the {\em SemEval16-Food} dataset and show the detailed results in Table \ref{tab:case}. 

The first two rows of Table \ref{tab:case} shows two cases where \textsf{HEF} successfully predicts the query's parent. We can see that the Pathfinder Score and the three Stopper Scores all contribute to the correct selection, which testifies the effectiveness of the Fitting Score design. 

The last two rows of Table \ref{tab:case} provide situations when \textsf{HEF} fails to select the correct parent. In the third row, {\em ``bourguignon''} is described as ``reduced red wine'', thus the model attaches it to the node {\em ``wine''}. However, {\em ``bourguignon''} is also a sauce for cooking beef. Such ambiguation consequently affects the meaning of a term by assigning an incorrect description, which hurts the model's performance. In the last row, although {\em ``hot fudge sauce''}'s description contains "chocolate sauce", the node {\em ``chocolate sauce''} still gets a low Current Score. In \textsf{HEF}, the design of Stopper Scores enables the model to self-correct the occasionally wrong Current Scores by assigning larger Forward Score from a node's parent and larger Backward Score from one of the node's children. However, since {\em ``chocolate sauce''} is a leaf node, its child's Backward Score is assigned to be the proportion of leaf nodes in the seed taxonomy, which is 0.07 in the {\em SemEval16-Food} dataset. This indicates that future work includes designing a more reasonable Backward Score function for leaf nodes to improve the model's robustness.

\section{Description Generation Algorithm}

Algorithm~\ref{alg:description} shows the description generation algorithm \texttt{descr($\cdot$)} used in \textsf{HEF}'s hypernymy detection module. \texttt{descr($\cdot$)} utilizes WordNet descriptions to generate domain-related term descriptions by dynamic programming. In this algorithm, \texttt{WordNetNounDescr($\cdot$)} means the set of a concept's noun descriptions from WordNet \cite{miller_wordnet_1995}, and  \texttt{CosSimilarity($t,n_{\mbox{root}}$)} means calculating the average token cosine similarity of word vectors between a candidate description $t$ and the surface name of a taxonomy's root term $n_{\mbox{root}}$.

\begin{algorithm}[htbp]
\caption{Description Generation Algorithm for the Hypernymy Detection Module}
\label{alg:description}
\begin{algorithmic}[1]
\Procedure{Descr}{$n$} \Comment{Input: term $n$}
    \State $N\leftarrow${\texttt{split($n$)}}
    \For{$i\leftarrow0,\cdots,\texttt{length($N$)}$}
        \State $S[i]=0$ \Comment{Initialize score array}
        \State $C[i]=0$ \Comment{Initialize splitting positions}
    \EndFor
    \For{$i\leftarrow0,\cdots,\texttt{length($N$)$-1$}$}
        \For{$j\leftarrow0,\cdots,i$}
            \If{\texttt{WordNetNounDescr($N[j:i+1]$)}$>0$}
                \State $s_{ij}=\left(i-j+1\right)^2+1$ \Comment{Prefer longer concepts}
            \Else
                \State $s_{ij}=1$
            \EndIf
            \If{$S[j]+s_{ij}>S[i+1]$}
                \State $S[i+1]\leftarrow S[j]+s_{ij}$ \Comment{Save max score}
                \State $C[i]=j$ \Comment{Save splitting position}
            \EndIf
        \EndFor
    \EndFor
    \State $D\leftarrow$``'' \Comment{Initialize description}
    \State $p\leftarrow\texttt{length(}N\texttt{)}$ \Comment{Generate split pointer}
    \While{$p\neq-1$}
        \State $D_{WN}=$\texttt{WordNetNounDescr($N\left[C[p]:p+1\right]$)}
        \If{{\texttt len($D_{WN}$)}$>0$} \Comment{Noun or noun phrase}
            \State $d\leftarrow\mathop{\arg\max}_{t\in D_{WN}}{ {\texttt{CosSimilarity(}}t,n_{\mbox{root}}{\texttt )}}$
        \Else  \Comment{Prep. or adj.}
            \State $d\leftarrow {\texttt join(}N[C[p]:p+1]\texttt{)}$
        \EndIf
        \State $D\leftarrow d+D$ \Comment{Put new description in the front}
        \State $p\leftarrow C[p]-1$ \Comment{Go to next split}
    \EndWhile
\EndProcedure
\end{algorithmic}
\end{algorithm}

\begin{table*}[t]
    \caption{Examples of \textsf{HEF}'s prediction, with detailed Fitting Score composition and comparison between the ground truth and the predicted parent. Scores in this table correspond to the node in the same tabular cell with the score.}
    \label{tab:case}
    \begin{tabularx}{\textwidth}{Q|PS|PS}
    \toprule
    \makecell[c]{Query ($q$)} & \makecell[c]{Ground Truth ($\hat{p}$)} & Scores & \makecell[c]{Prediction ($p$)} & Scores \\
    \midrule
    \multirow{3}{\hsize}{$q$: \textbf{paddy} is rice in the husk either gathered or still in the field} & \multirow{2}{\hsize}{$\hat{p}$: \textbf{rice} is grains used as food either unpolished or more often polished} & $S_p=0.9997$ & \multirow{2}{\hsize}{$p$: \textbf{rice} is grains used as food either unpolished or more often polished} & $S_p=0.9997$ \\
    & & $S_c=0.4599$ & & $S_c=0.4599$ \\\cmidrule{2-5}
    & ${\rm parent}(\hat{p})$: \textbf{starches} is a commercial preparation of starch that is used to stiffen textile fabrics in laundering & $S_f=0.9755$ & ${\rm parent}(p)$: \textbf{starches} is a commercial preparation of starch that is used to stiffen textile fabrics in laundering & $S_f=0.9755$ \\\cmidrule{2-5}
    \makecell[c]{$F(\hat{p},q)=0.4483$} & \multirow{2}{\hsize}{$c_{\hat{p}}^*$: \textbf{white rice} is having husk or outer brown layers removed} & \multirow{2}{*}{$S_b=0.9995$} & \multirow{2}{\hsize}{$c_p^*$: \textbf{white rice} is having husk or outer brown layers removed} & \multirow{2}{*}{$S_b=0.9995$} \\
    \makecell[c]{$\hat{p}$'s Ranking: 1} & & & & \\
    \midrule
    
    \multirow{3}{\hsize}{$q$: \textbf{fish meal} is ground dried fish used as fertilizer and as feed for domestic livestock} & \multirow{2}{\hsize}{$\hat{p}$: \textbf{feed} is food for domestic livestock} & $S_p=0.9993$ & \multirow{2}{\hsize}{$p$: \textbf{feed} is food for domestic livestock} & $S_p=0.9993$ \\
    & & $S_c=0.3169$ & & $S_c=0.3169$ \\\cmidrule{2-5}
    & ${\rm parent}(\hat{p})$: \textbf{food} is any substance that can be metabolized by an animal to give energy and build tissue & $S_f=0.9984$ & ${\rm parent}(p)$: \textbf{food} is any substance that can be metabolized by an animal to give energy and build tissue & $S_f=0.9984$ \\\cmidrule{2-5}
    \makecell[c]{$F(\hat{p},q)=0.3158$} & \multirow{2}{\hsize}{$c_{\hat{p}}^*$: \textbf{mash} is mixture of ground animal feeds} & \multirow{2}{*}{$S_b=0.9988$} & \multirow{2}{\hsize}{$c_p^*$: \textbf{mash} is mixture of ground animal feeds} & \multirow{2}{*}{$S_b=0.9988$} \\
    \makecell[c]{$\hat{p}$'s Ranking: 1} & & & & \\
    \midrule
    
    \multirow{3}{\hsize}{$q$: \textbf{bourguignon} is reduced red wine with onions and parsley and thyme and butter} & \multirow{2}{\hsize}{$\hat{p}$: \textbf{sauce} is flavorful relish or dressing or topping served as an accompaniment$\cdots$} & $S_p=0.0002$ & \multirow{2}{\hsize}{$p$: \textbf{wine} is a red as dark as red wine} & $S_p=0.9997$ \\
    & & $S_c=0.0001$ & & $S_c=0.1399$ \\\cmidrule{2-5}
    & ${\rm parent}(\hat{p})$: \textbf{condiment} is a preparation (a sauce or relish or spice) to enhance flavor or enjoyment & $S_f=0.0004$ & ${\rm parent}(p)$: \textbf{alcohol} is any of a series of volatile hydroxyl compounds that are made from hydrocarbons by distillation & $S_f=0.9812$ \\\cmidrule{2-5}
    \makecell[c]{$F(\hat{p},q)=1e-11$} & \multirow{2}{\hsize}{$c_{\hat{p}}^*$: \textbf{bercy} is butter creamed with white wine and shallots and parsley} & \multirow{2}{*}{$S_b=0.9997$} & \multirow{2}{\hsize}{$c_p^*$: \textbf{red wine} is wine having a red color derived from skins of dark-colored grapes} & \multirow{2}{*}{$S_b=0.8784$} \\
    \makecell[c]{$\hat{p}$'s Ranking: 328} & & & & \\
    \midrule
    
    \multirow{3}{\hsize}{$q$: \textbf{hot fudge sauce} is hot thick chocolate sauce served hot} & \multirow{2}{\hsize}{$\hat{p}$: \textbf{chocolate sauce} is sauce made with unsweetened chocolate or cocoa$\cdots$} & $S_p=0.9471$ & \multirow{2}{\hsize}{$p$: \textbf{sauce} is flavorful relish or dressing or topping served as an accompaniment$\cdots$} & $S_p=0.9995$ \\
    & & $S_c=9e-5$ & & $S_c=0.0172$ \\\cmidrule{2-5}
    & ${\rm parent}(\hat{p})$: \textbf{sauce} is flavorful relish or dressing or topping served as an accompaniment$\cdots$ & $S_f=0.9617$ & ${\rm parent}(p)$: \textbf{condiment} is a preparation (a sauce or relish or spice) to enhance flavor or enjoyment & $S_f=0.9888$ \\\cmidrule{2-5}
    \makecell[c]{$F(\hat{p},q)=6e-6$} & \multirow{2}{\hsize}{$c_{\hat{p}}^*$: \textbf{None}} & \multirow{2}{*}{$S_b=0.0700$} & \multirow{2}{\hsize}{$c_p^*$: \textbf{lyonnaise sauce} is brown sauce with sauteed chopped onions and parsley$\cdots$} & \multirow{2}{*}{$S_b=0.9995$} \\
    \makecell[c]{$\hat{p}$'s Ranking: 20} & & & & \\
    \bottomrule
  \end{tabularx}
\end{table*}

\end{document}